\documentclass[letterpaper]{article}
\usepackage{calc,amsmath,amssymb,amsfonts}
\usepackage[LGR,T1]{fontenc}
\usepackage[greek,english]{babel}
\usepackage{xcolor,longfbox,fancyhdr}
\usepackage[vmargin=0.5in,hmargin=1in,includeheadfoot,head=0.5in,headsep=0.4602in,foot=0.5in,footskip=0.9602in]{geometry}
\usepackage{enumitem,hyperref}
\hypersetup{colorlinks=true,allcolors=blue,pdfauthor=Ho-Joon Lee; Ph.D.}
\usepackage[pdftex]{graphicx}
\makeatletter\newdimen\@tempdimd\makeatother
\newcommand\textstyleListLabelccxxxi[1]{\textrm{#1}}
\newcommand\textstyleListLabelccxli[1]{\textrm{#1}}
\newcommand\textstyleListLabelcxcv[1]{\textrm{#1}}
\newcommand\textstyleListLabelcciv[1]{\textrm{#1}}
\fancypagestyle{Standard}{\fancyhf{}
  \fancyhead[R]{}
  \fancyfoot[L]{\thepage{}}

  \renewcommand\thepage{\arabic{page}}
}
\author{Ho-Joon Lee, Ph.D.}
\date{2024-09-27}

\begin{document}
\clearpage
\pagestyle{Standard}
\setcounter{page}{1}

\section[Improved Prediction Of Ligand{}-Protein Binding Affinities By Meta{}-Modeling]{Improved Prediction Of
Ligand-Protein Binding Affinities By Meta-Modeling}
\bigskip

Ho-Joon Lee\textsuperscript{1,2,*\#}, Prashant S. Emani\textsuperscript{3,*\#} and Mark B.
Gerstein\textsuperscript{3,4,5,6,7\#}
\bigskip

\textsuperscript{1}Dept. of Genetics and \textsuperscript{2}Yale Center for Genome Analysis, Yale University, New Haven,
CT 06510, USA\par
\textsuperscript{3}Dept. of Molecular Biophysics \& Biochemistry, Yale University, New Haven, CT 06520, USA\par
\textsuperscript{4}Program in Computational Biology \& Bioinformatics, \textsuperscript{5}Dept. of Computer Science,
\textsuperscript{6}Dept. of Statistics \& Data Science, and \textsuperscript{7}Dept. of Biomedical Informatics \& Data
Science, Yale University, New Haven, CT 06520, USA\par
\bigskip

* Equal contributions\par
\# Correspondence: HL, \href{mailto:ho-joon.lee@yale.edu}{\textcolor{blue}{ho-joon.lee@yale.edu}}; PSE,
\href{mailto:prashant.emani@yale.edu}{\textcolor{blue}{prashant.emani@yale.edu}}; MBG,
\href{mailto:pi@gersteinlab.org}{\textcolor[HTML]{1155CC}{pi@gersteinlab.org}}


\clearpage
\section{Abstract}
\textcolor{black}{The accurate screening of candidate drug ligands against target proteins through computational
approaches is of prime interest to drug development efforts. Such virtual screening depends in part on methods to
predict the binding affinity between ligands and proteins. Many computational models for binding affinity prediction
have been developed, but with varying results across targets. }Given that ensembling or meta-modeling approaches have
shown great promise in reducing model-specific biases\textcolor{black}{, we develop a framework to integrate published
force-field-based empirical docking and sequence-based deep learning models. In building this framework, we evaluate
many combinations of individual base models, training databases, and several meta-modeling approaches. We show that
many of our meta-models significantly improve affinity predictions over base models. Our best meta-models achieve
comparable performance to state-of-the-art deep learning tools exclusively based on 3D structures, }while allowing for
improved database scalability and flexibility through the explicit inclusion of features such as physicochemical
properties or molecular descriptors\textcolor{black}{. We further demonstrate improved generalization capability by our
models using a large-scale benchmark of affinity prediction as well as a virtual screening application benchmark.
Overall, we demonstrate that diverse modeling approaches can be ensembled together to gain meaningful improvement in
binding affinity prediction.}


\clearpage
\section{Introduction}
Molecular interactions such as protein-protein or ligand-protein interactions are fundamental processes in biology as
well as biophysics and biochemistry. Knowledge of interaction energy or binding affinity is essential in understanding
biological functions and biomedical applications. In drug discovery or development, accurate prediction of binding
affinity of a drug ligand bound to a target molecule (typically a protein) is of central importance in identifying the
most stable structure of a drug-target complex, i.e., the best 3-dimensional pose (geometry) of a drug in the binding
pocket of a target molecule, known as molecular docking. Along with development of molecular docking algorithms, there
have been extensive computational studies with decades of efforts to achieve better prediction of ligand-protein
binding affinity with applications to high throughput drug screening or virtual screening \textsuperscript{1-10}. This
has been achieved by various scoring functions that represent the binding free energy or stability of a ligand-protein
complex and select the best docking pose. Docking tools typically generate multiple candidate poses of a ligand and
rank them by binding affinity scores. Therefore, accurate prediction of binding affinity is critical in drug discovery
and development. Scoring functions were developed traditionally by \textit{ab initio} physical force field-based,
knowledge-based, or empirical methods, or more recently by advanced machine learning or deep learning approaches.
\bigskip

The predictive power of traditional or classical scoring functions is known to be very limited due to rigid or naive
underlying assumptions \textsuperscript{1, 5}, e.g. reliance on linear regression \textsuperscript{11, 12} or
over-simplification or ignorance of desolvation and entropy contributions because of computational cost
\textsuperscript{13-15}. In addition, previous evaluation studies of more than a dozen scoring functions reported that
no scoring function was consistently better than others in all conducted tests, showing both strengths and weaknesses
for different purposes \textsuperscript{16-18}. For example, the best scoring function in terms of a prediction
correlation with experimental binding affinities was not necessarily best in terms of predicting correct ligand poses
(i.e., not the lowest prediction errors). Despite those limitations, classical scoring functions (especially empirical
ones) have been widely used and implemented in several software tools such as Autodock Vina \textsuperscript{19} or
\textit{smina} \textsuperscript{20}. Furthermore, consensus scoring strategies from multiple scoring functions have
shown improved performances \textsuperscript{16, 21-24}.
\bigskip

In continued efforts to improve prediction accuracy, advanced machine or statistical learning-based regression methods
such as Random Forests, support vector machines, or neural networks have been successfully used to develop alternative
scoring functions by leveraging large-scale databases such as PDBbind \textsuperscript{25} for more diverse
physicochemical and geometric features of ligand-protein complexes and their experimental binding affinities
\textsuperscript{1, 3, 11, 26-28}. Improved performance of machine learning (ML) methods is mainly due to their ability
to leverage big data and engineered features to capture non-linear or non-additive relationships in binding or docking
that are likely to be missed by classical scoring functions \textsuperscript{29}. Furthermore, with the groundbreaking
advancement of deep learning (DL) in computer science, its applications in biomedicine, including protein structure
prediction, binding affinity prediction, and drug discovery, have been very successful and promising
\textsuperscript{1, 27, 30-39}. Deep learning methods for binding affinity prediction are advanced regression methods
that typically generate sophisticated abstract representations or embeddings for ligands, proteins, or ligand-protein
complexes via various neural network architectures to learn hidden patterns. Inspired by the breakthroughs in computer
vision \textsuperscript{40, 41}, various DL models based on convolutional neural networks (CNNs) have been developed
using 3-dimensional (3D) structural information of ligand-protein complexes with high prediction accuracies superior to
those advanced ML models mentioned above \textsuperscript{42-50}. In addition, methods based on graph neural networks
have been successful as well \textsuperscript{42, 47, 51-56}. On the other hand, DL methods with ligand and protein
sequences alone, i.e., without structural information, have also shown comparable performances \textsuperscript{57,
58}. In general, DL architectures are highly diverse with a number of hyperparameters such as the network depth and
width, loss and activation functions, and a learning rate, which makes model optimization very challenging due to high
computational cost. More importantly, fundamental issues with DL approaches have been interpretability or
explainability of their non-intuitive excellent performances \textsuperscript{59-63}. On another note, mathematically
advanced methods such as differential geometry and/or algebraic topology-based analyses in combination with ML or DL
regression have been also developed showing notably improved performances in ligand-protein binding affinity prediction
\textsuperscript{64-71}, as well as physics-based hybrid models \textsuperscript{72, 73}.
\bigskip

In this work, we aim to improve ligand-protein binding affinity prediction by combining both classical scoring
functions, which are built upon physical force fields, and sequence-based DL models into a single integrative framework
of meta-models (also called ensemble, fusion, or hybrid models). Meta-models are a type of ensemble learning by
combining diverse algorithms and have been successful in machine or statistical learning for improved results with more
robustness and less bias than individual algorithms (known as stacked generalization or super learning)
\textsuperscript{74, 75}. Meta-models, termed \textit{blending} in associated publications, were also very successful
in addressing the Netflix Prize challenges \textsuperscript{76-79}. In predicting ligand-protein binding affinity,
ensemble models of multiple 3D CNNs have been shown to outperform single CNNs and previous models \textsuperscript{50}.
On the other hand, theory-driven physical models and data-driven ML or statistical models are complementary to each
other. The former offers physical interpretability and constraints, whereas the latter is better at complex pattern
discovery from massive data. For this reason, hybrid approaches have been favored in recent years \textsuperscript{72,
73, 80-83}. In general, the intuition behind meta-modeling is that different solutions to a problem could be modeling
non-overlapping aspects of the ground truth. These different solutions may be trained on data with their own systematic
biases, or might have architectures that access some limited subspace of the possible model space. When combined, an
ensemble of such solutions might serve to better represent the ground truth. We have designed a modeling framework that
allows the inclusion of structure-based classical scoring functions and sequence-based DL models, as well as explicit
inclusion of covariates such as physicochemical features of ligand, proteins, or ligand-protein complexes
(\textbf{Figure 1 and Supplementary Figure S1}). In this study, we first focus on 2 empirical scoring functions (ESFs),
SMINA and Vinardo (implemented in \textit{smina}) \textsuperscript{20} together with consensus docking strategies based
on experimental or predicted ligand poses. For DL models, we use sequence-based DL models of 10 different architectures
with multiple training strategies, employing an open-source DL library for drug-target interaction prediction called
DeepPurpose \textsuperscript{57} and two ligand-protein binding databases, BindingDB \textsuperscript{84} and PDBbind.
We investigate both pre-trained models from the DeepPurpose library and our own fine-tuned or \textit{de
novo}{}-trained models using PDBbind, which is the main data used by the ESF models. In addition, by merging all
prediction scores from repeated cross-validations for all DL models under study, we investigate top principal
components from principal component analysis (PCA) as additional prediction scores. Finally, by taking all prediction
scores from both ESF and DL models as well as molecular weights of ligands, we build meta-models using 4 machine
learning algorithms and 11 feature sets as our final tools (44 in total) for ligand-protein binding affinity
prediction. 
\bigskip

Using 3 diverse benchmark evaluations of CASF-2016 \textsuperscript{7}, the large-scale PDBbind v2020 general set, and a
virtual screening dataset of LIT-PCBA \textsuperscript{85}, we demonstrate that our meta-models outperform individual
base models and exhibit competitive performance to more sophisticated structure-based DL models, serving as a
potentially useful general framework for continual future improvement of ligand-protein binding affinity prediction.
The advantages offered by such an approach over existing methods include: (1) scalability by utilizing sequence-based
databases for DL models without the need for 3D structures; and (2) flexibility by allowing different base models and
physicochemical or molecular properties to be incorporated into meta-models. We believe that such an extensive
exploration of these aspects of binding affinity prediction makes a meaningful contribution to the discussion of model
generalization.

\clearpage\section{Methods}
\subsection{Dataset selection}
\textit{BindingDB}\par
\textcolor{black}{We used the BindingDB ver. 2020m2 database (}\url{http://www.bindingdb.org}\textcolor{black}{) for the
purpose of DL model training. A total of 66,444 ligand-protein complexes from the Ligand-Target-Affinity dataset were
filtered with Kd {\textless}= 0.01M, valid PubChem IDs or InChI keys for ligands, and valid UniProt IDs for proteins.}
\bigskip

{\raggedright\textit{PDBbind}\par}
\textcolor{black}{We downloaded the PDBbind v2020 }\textit{\textcolor{black}{refined}}\textcolor{black}{
}\textit{\textcolor{black}{set}}\textcolor{black}{ from }\url{http://www.pdbbind.org.cn/}\textcolor{black}{. The
dataset consisted of 5,316 target protein-drug ligand complexes. The }\textit{\textcolor{black}{refined
set}}\textcolor{black}{ (labeled RefinedSet) has been filtered by the PD}Bbind creators \textcolor{black}{from a larger
dataset by the removal of complexes that have “obvious problems in 3D structure, binding data or other aspects”
}\textcolor{black}{\textsuperscript{86}}\textcolor{black}{. Of these, 285 ligand-protein complexes are in the so-called
}\textit{\textcolor{black}{core}}\textcolor{black}{ set (labeled CoreSet), which are high-quality data that are used as
the primary test set for the Comparative Assessment of Scoring Functions (CASF) benchmark
}\textcolor{black}{\textsuperscript{7}}\textcolor{black}{. This CoreSet was separated from the remainder of the PDBbind
RefinedSet to serve as an external test set (i.e., a benchmark set) for evaluating model performance. Thus, we
considered a training set of 5,031 ligand-protein complexes and an external test set of 285 complexes. As 266 of 285
affinities were properly predicted upon running the docking tools (see below), we took those 266 complexes as our
primary benchmark set. Binding affinity values we used in this study were the natural logarithms of Kd or Ki values. We
did not distinguish Kd and Ki from each other.}

\subsection{Molecular docking}
\textit{Implementation of ligand docking}\par
We employ the \textit{smina} implementation \textsuperscript{20} of the AutoDock Vina method \textsuperscript{19} for
the docking component. The \textit{smina} implementation is designed to make AutoDock easier to use, with larger
support for ligand molecular formats and improved minimization algorithms. We used both the default SMINA scoring
function and another previously published scoring function, Vinardo \textsuperscript{87}, provided in \textit{smina}.
We note that the italicized \textit{smina} describes the program while the capitalized SMINA describes the scoring
function. In the meta-models below, we refer to this class of empirical docking methods as E. The tandem use of both
scoring functions is meant to address possible idiosyncrasies in the parameterizations of the scoring functions.
\textit{smina} is run as a command-line program using a receptor structure in PDBQT format and a ligand in 3D SDF
format. Our in-house Python script includes steps for the preparation of the receptor structures, the creation of the
configuration file, and the running of \textit{smina} (\textbf{Supporting Information}). The output for each scoring
function consists of the binding affinities for the 9 (the default option) best docking poses. Seven complexes with
multiple ligands (whose names were separated with “\&” or “/”) in the PDBbind dataset were not docked as there was some
ambiguity as to which ligand to consider in the docking. These complexes were excluded from further analyses.
\bigskip

{\raggedright\textit{Processing of docking scores}\par}
Previous studies have demonstrated the value of (a) filtering the docking poses output by empirical docking programs
based on the degree to which they agree with experimental scores \textsuperscript{20} and (b) comparing poses from two
or more docking methods, in so-called “consensus” approaches, and filtering based on the degree of agreement to yield a
higher quality output \textsuperscript{23, 24, 88}. Accordingly, we carried out preprocessing of the docking scores by
filtering out certain complexes based on the quality of the associated poses assessed by these two criteria. For each
complex out of the available 9 docking poses, we calculated the root-mean-square deviation (RMSD) between docking poses
in two different ways (described below), and utilized the RMSD value as a cutoff in determining the best set of binding
affinities to choose in the final dataset to the meta-models. The goal was to determine if pose quality, either
relative to an available experimental gold-standard or relative to the results of another scoring function, would
impact both the matching of the predicted binding affinity as well as the downstream prediction process.

We choose to implement the symmetric RMSD employed by Trott and Olson \textsuperscript{19} using
Pymol$\text{\textgreek{’}}$s Python API. In brief, this RMSD is calculated between two different ligand structures as 
$\mathit{RMSD}_{\mathit{ab}}=\mathit{max}\left(R_{a\vee b},R_{b\vee a}\right)$ \ where  $R_{a\vee b}$ \ describes an
asymmetric measure of distance: 

\begin{equation*}
R_{a\vee b}=\sqrt{\frac 1 N\sum _i\mathit{min}_jr_{\mathit{ij}}^2}
\end{equation*}
where the summation is over all  $N$ \ heavy atoms in structure  $a$, the minimum is over all atoms ( $j$) in structure 
$b$ \ of the same element type as atom  $i$ \ in structure  $a$. 

We calculate  $\mathit{RMSD}_{\mathit{ab}}$ \ in two ways, depending on the choice of docking poses: (1) Experimental
pose-RMSD filtering, where the RMSD is calculated for a ligand pose predicted by a docking method relative to the
experimental structure of the ligand (when available); and (2) Consensus pose-RMSD filtering, where the RMSD is
calculated between the docking poses for the same ligand-target pair as predicted by two different docking scoring
methods, Vinardo and SMINA. Details of these filtering approaches are provided in \textbf{Supporting Information}. We
also chose a cutoff of 3 Angstroms as representing a reasonably good quality match between poses based on our manual
inspection of RMSD distributions (\textbf{Supplementary Figure S2A}). This is a more lenient cutoff relative to other
analyses \textsuperscript{20, 23} that suggest a 2-Angstrom cutoff. Our approach is a compromise between the need for
higher quality matches, and the need to have sufficient numbers of data points to train our models as stricter cutoffs
lead to smaller dataset sizes.

An important aspect of these parallel approaches to applying the RMSD filters on the data is that the scores extracted
from the empirical docking runs are different when considering the Experimental RMSD filter and the Consensus RMSD
filter. This is even the case when considering the unfiltered dataset, where the set of complexes are identical, but
the choice of poses used to extract the best docking scores differ between the two RMSD filter approaches. Another
aspect we note is that experimental structures are not necessarily available for protein-ligand complexes used in many
downstream use cases. It is therefore not our intention that users will apply these filters on the prediction dataset.
Rather, we utilize the filters on our training data to produce what is hopefully a more informative dataset for the
meta-model training. Our evaluations of the relative merits of different filters on the training set are provided in
Results. Finally, we note that, in the naming of the model files, we use the shorthand “VvS” to indicate the Consensus
RMSD filtering as the comparison involves Vinardo versus SMINA poses.

\subsection{Deep Learning}
\textit{DeepPurpose library}\par
\textcolor{black}{We used the Python library, DeepPurpose, for deep learning of binding affinities of ligand-target
complexes }\textcolor{black}{\textsuperscript{57}}\textcolor{black}{. It offers a framework for various models that
combine ligand-encoding and protein-encoding neural networks. We modified and customized the version 0.0.1 of the
library (as of July 24, 2020) for development of our work, while all 6 pre-trained models with BindingDB-2020m2-Kd were
downloaded from the version 0.1.5 (as of June 29, 2022). The 6 pre-trained models are CNN-CNN, Daylight-AAC,
Morgan-AAC, Morgan-CNN, MPNN-CNN, and Transformer-CNN. We refer to this class of BindingDB pre-trained models as D1.}
\bigskip

{\raggedright\textit{Training of 12 DL models from the DeepPurpose library using BindingDB}}\par
\textcolor{black}{The DeepPurpose library offers various embedding methods for both ligands and proteins to build
regression models to predict binding affinities of ligand-protein complexes. After initial pilot experiments, we chose
the following 12 models of ligand-protein embedding methods for downstream analyses: CNN-CNN, Daylight-AAC, Morgan-AAC,
Morgan-CNN, MPNN-CNN, Transformer-CNN, MPNN-Quasi-seq, CNN/RNN-CNN/RNN, CNN-AAC, MPNN-Transformer, MPNN-PseudoAAC, MPNN-AAC, MPNN-CNN/RNN, Transformer-Quasi-seq, and Transformer-Transformer. We applied the 12 models to BindingDB Kd data, which were used by the DeepPurpose library. Their 6 pre-trained models mentioned above were reported to achieve performances of mean 5-fold cross-validated MSE {\textless} 0.6 or Pearson correlation coefficient (PCC) {\textless}
0.8 }\textcolor{black}{\textsuperscript{57}}\textcolor{black}{. Given the published results of their best models, we
re-trained or }\textit{\textcolor{black}{de novo}}\textcolor{black}{{}-trained all 12 models from scratch to either
reproduce their best models or identify more models with similar performances for our downstream model development. We
used all default hyperparameters for each model in DeepPurpose and did not perform cross validation given the target
performance of MSE = 0.6 or PCC = 0.8. The data split for training, validation, and testing was 70\% (46,511 cpx), 10\%
(6,644 cpx), and 20\% (13,289 cpx), respectively, with random seed = 1. Those top models selected with the target
performance are considered as crude or low-confidence deep learning models, being a starting point in our entire
pipeline for the sake of }\textcolor{black}{computational efficiency given limited resources. We refer to this class of
BindingDB de novo-trained models as D2.}
\bigskip

{\raggedright\textit{Training of DL models using PDBbind}}\par
\textcolor{black}{We also trained 10 DL models using the PDBbind RefinedSet binding affinity data (Kd/Ki) that contain
ligand-target structural data and were used by the docking tools. The 10 models come from the 6 pre-trained models and
the top 6 }\textit{\textcolor{black}{de novo}}\textcolor{black}{{}-trained models selected from above (two models in
common). This is to compare with the BindingDB-trained DL models. We excluded the CASF2016 CoreSet within the PDBbind
RefinedSet, which we use for external validation, and performed 10x repeated 5-fold cross validation with random seed =
1701. We refer to this class of PDBbind-trained models as D3.}
\bigskip

{\raggedright\textit{Fine-tuning of the BDB-trained models using PDBbind}}\par
\textcolor{black}{In order to leverage the respectable performance of the BindingDB-trained models (D1 and D2), we
finetuned them using the PDBbind data excluding the }CoreS\textcolor{black}{et (referred as D1F and D2F, respectively).
To increase analysis robustness, we performed iterative 5-fold CV with 10 iterations, i.e., 50 validations (using the
scikit-learn function in Python, RepeatedKFold or KFold, with random\_state=1701) for each of the 6 D1F, 6 D2F, and 10
D3 models to yield 300, 300, 500 model variants or predictions, respectively. We also performed PCA on those
predictions for the 3 model classes (referred as D1FP, D2FP, D3P, respectively) as well as on the merged 1,100
predictions (referred as DAP) for ML meta-models below for dimensionality reduction of the meta-features (predictors).}

\subsection[Machine Learning Meta{}-Models]{Machine Learning Meta-Models}
The fact that we have two independent approaches of quantifying the binding affinities (i.e. docking and deep learning
tools) means that we potentially have complementary sources of information in the data. The docking tools are based on
empirical scoring functions trained to match experimental structural data, while the deep learning tools capture
information on the sequences of ligands and proteins. The exact partition of dual sequence/structure impacts on the
results from these two approaches is not entirely clear, but the expectation is that there is likely non-overlapping
information contained in the predictions of these approaches. Accordingly, we have sought to exploit any complementary
information by feeding the docking and deep learning scores into a set of linear and non-linear machine learning models
to gain improvements in prediction of binding affinities. We term these machine learning models as \textit{meta-models}
since they incorporate predictions from independent tools as base models. We used linear regression, ElasticNet, and
LASSO algorithms for linear meta-models and the XGBoost algorithm for a non-linear meta-model implemented in
\textit{scikit-learn} in Python. Hyperparameter optimization was done along with multiple 5-fold cross validations.
Details are given in \textbf{Supporting Information}. 
\bigskip

The features for the meta-models are SMINA and Vinardo predictions with different RMSD filtering cutoffs, ligand
molecular weights (MWs), and DL predictions. We experimented with various combinations of those features: (1) SMINA +
Vinardo (with/without the filters) (2) SMINA + Vinardo (with/without the filters) + MW (3) SMINA + Vinardo + MW + DL
mean predictions (BindingDB or PDBbind) and (4) SMINA + Vinardo + MW + DL PCA projections (BindingDB or PDBbind). For
the BindingDB-trained scores, we used the top 6 models as described above. For the DL models fine-tuned on the PDBbind
dataset, we take the mean of the 50 CV predictions for each of the 6 models (i.e., 6 meta-model features). For the PCA
of all DL scores, we also ran an optimization over the number of principal components (PCs) to include as features. We
allowed for between 1 and 22 top principal components to be included (i.e., PC 1, PCs 1-2, PCs 1-3, … PCs 1-22), and
the selection of the best model among the 22 options was done by maximizing the Pearson correlation for each of the 4
meta-models separately using 20\% of the training data as a leave-out validation set for hyperparameter optimization.
We chose to optimize up to the top 22 PCs to match the number of models used in all the deep learning runs in D1, D2
and D3 combined. 
\bigskip

Each of the optimized meta-models generated above is tested against the CoreSet using the relevant meta-model-specific
feature sets. For all meta-models, the Pearson correlation, the Spearman rank correlation, the mean-square error (MSE),
and the root-mean-square error (RMSE) for the CoreSet predictions were calculated and reported for model evaluation. A
detailed description of these meta-model features is shown in \textbf{Table 1}.

\subsection{SwissADME and UniProt Analysis}
\textcolor{black}{To understand the contributions of ligand properties to benchmark performance of all the models, we
used the web tool }\textit{\textcolor{black}{SwissADME}}\textcolor{black}{
}\textcolor{black}{\textsuperscript{89}}\textcolor{black}{ to extract the properties of the PDBbind CoreSet ligands
from their SMILES strings. Also, we investigated the SwissADME features that might be associated with better or worse
performance for subsets of complexes by the different groups of tools }(docking, DL, and meta-models). To evaluate the
contributions of protein features, we extracted the relevant subsets of PDBbind CoreSet PDB IDs from our analyses and
fed them into the “Retrieve/ID mapping” tool of the UniProt database \textsuperscript{90}. We studied the resulting
protein properties (e.g. pathway information) for over- or under-representation in our predictions. Details are
provided in \textbf{Supporting Information}.

\subsection{Comparison to other tools}
We compared our models to 4 recently published structure-based tools. For fair comparisons, we re-trained those tools
using our training dataset as long as there were no technical difficulties. Kyro et al. developed a new structure-based
DL ensemble or meta model, HAC-Net \textsuperscript{47}, by combining MP-GNN (message passing graph neural networks)
and 3D-CNN (convolutional neural networks) trained on the PDBbind-2020 general set of {\textgreater}18,800
ligand-protein complexes. For a fairer, but limited, comparison, we re-trained HAC-Net (v1.3.1) with our training set of the
PDBbind-2020 RefinedSet{\textbackslash}CoreSet by subsetting from their train/validation/test data splits (4,942/82/265
out of 18,818/300/290 complexes from the general set, respectively). Jones et al. developed meta or fusion models, FAST
\textsuperscript{42}, by combining 3D-CNN and SG-CNN (spatial graph CNN). Due to technical difficulties, we were able
to only run their pre-trained SG-CNN model trained on the PDBbind-2016 general set excluding the CoreSet. KDeep
\textsuperscript{46} was one of the earlier structure-based DL models with 3D-CNN to predict ligand-protein binding
affinity. The KDeep model available on the public web-server \textsuperscript{91} could be re-trained with limited data
size (KdeepTrainer, https://playmolecule.com/KdeepTrainer/). While we were not able to use our full training data due
to the limitation, we were able to re-train the KDeep model for 250 epochs with each of 3 training sets of 1,800 random
complexes sampled from our training set of the RefinedSet{\textbackslash}CoreSet. PerSpect ML \textsuperscript{69} is
one of the best performing tools in recent years based on topological data analysis and spectral theory for multiscale
molecular featurization, which may generate more than 100,000 features from 3D structures or binding poses for the
gradient boosting tree-based regression. Due to incomplete code availability, it was not possible to reproduce or
implement PerSpect ML for fair comparisons with the data we used. We compare our results to their published results
with that caveat.

\subsection{GeneralSet Benchmark}
To assess model generalizability on a large scale, we performed an external validation of binding affinities for
complexes in the PDBbind v2020 GeneralSet as another benchmark in addition to the CoreSet. The full GeneralSet consists
of 19,443 complexes and is not as carefully curated as the RefinedSet. Hence, we filtered complexes according to the
following rules:

\begin{enumerate}[series=listWWNumxxix,label=\arabic*.,ref=\arabic*]
\item Filter out all complexes belonging to the RefinedSet and CoreSet.
\item Remove all complexes assayed by NMR.
\item Remove all complexes with an IC50 dissociation measure (as opposed to Kd or Ki).
\item Only include complexes whose PDB structures were published after the year 2000.
\item Finally, not all complexes had successful docking calculations for both Vinardo and SMINA. We left out complexes
with any predicted binding affinity of zero.
\end{enumerate}
The resulting subset included 5,261 complexes, which is larger than our training set of 5,031 complexes from the
RefinedSet. Therefore, it is a more challenging benchmark than the CoreSet. We ran the docking and our model
predictions using the same settings as for the CoreSet.

\subsection{Ranking Benchmark}
To test the possibility of certain tools performing the prediction task with differing degrees of accuracy for certain
proteins or even families of proteins, we devised a virtual screening test by selecting 3 most frequently targeted
proteins by multiple ligands from the PDBbind GeneralSet excluding the RefinedSet we used for model training. We
considered only those proteins in GeneralSet whose sequence similarity to those in RefinedSet was found to be less than
30\% by the NCBI Blastp package \textsuperscript{92}. We also filtered out complexes in the GeneralSet that had only
IC50 experimental measurements for consistency with this study. The resulting top 3 proteins are O60341, O15151, and
P0C6U8 (UniProt Accession IDs), which are complexed with 20, 8, and 7 ligands, respectively. The resolution of the 35
complexes ranges from 1.33 to 3.50 Angstroms with mean of 2.58 and standard deviation of 0.69.

\subsection{Virtual Screening}
For a more general applicability, we also performed a virtual screening using an external benchmark dataset, LIT-PCBA,
of active and inactive ligands for 15 target proteins \textsuperscript{85}. Although our models were neither explicitly
designed nor trained to distinguish active and inactive ligands from each other (i.e., binary classification for
functional activity), our assumption was that active ligands tend to have higher binding affinities (i.e., lower Kd/Ki)
for target proteins than inactive ligands. This assumption is weak because there is an unclear causal relationship
between binding affinity (Kd, Ki) and functional activity (IC50, EC50). We also note that we used ligand-protein
complexes with Kd or Ki from both BindingDB and PDBbind for consistency. LIT-PCBA is a recently developed virtual
screening dataset which was reported to be less biased than previous datasets. It is based on PubChem BioAssays which
includes 4 potency types of IC50, EC50, Kd, and Ki. We point out that LIT-PCBA did not distinguish the 4 potency types
from each other. As a proof-of-concept, with the aforementioned caveat in mind, we randomly selected active and
inactive ligands maintaining the ratio 1:2 (number of active ligands:number of inactive ligands) for each of the 15
proteins in LIT-PCBA to predict their binding affinities by our models. Specifically, if the number of active ligands
was {\textless} 100, we included them all; if greater than or equal to 100, we randomly sampled 100 active ligands. The
inactive ligand numbers were then selected to maintain the aforementioned 1:2 ratio. This gave us a total of 1,050
active ligand-target protein pairs and 2,100 inactive ligand-target protein pairs. The active and inactive ligands were
chosen from the combined training and validation sets in LIT-PCBA. We generated three-dimensional SDF files for each of
the ligands from the SMILES strings. We also randomly selected only one of the PDB structures (in mol2 format) for each
target protein in this virtual screening exercise (further details in \textbf{Supporting Information}). The evaluation
was done in 4 ways: (1) comparisons of predicted binding affinity distributions between active and inactive ligands for
each protein with Welch$\text{\textgreek{’}}$s t-test and the Mann-Whitney U test for significant differences; (2) top5
recall: an enrichment or fraction of active ligands (hits) in the top 5 highest-affinity ligands predicted for each
protein; (3) similarly, top10 recall for hits in the top 10 ligands; and (4) precision for hits in the number of the
top ligands equal to that of all active ligands for each protein.

\clearpage\section{Results}
\subsection{Docking tools}
We ran \textit{smina} with two different scoring functions, SMINA and Vinardo (model class E), on the PDBbind dataset.
We distinguish the PDBbind “RefinedSet” excluding the “CoreSet” (designated as “RefinedSet{\textbackslash}CoreSet”)
from the “CoreSet” which is used as the external validation (benchmark) set in this study. Additionally, as described
in the Methods, we considered two formulations of the docking pose-based RMSD for the 9 poses generated by the docking
tools for each protein-ligand complex: an “Experimental” RMSD relative to the experimental pose, and a “Consensus”
RMSD. We indicate the unfiltered dataset by RMSD {\textless} 101 Angstroms, as all structures will satisfy this
condition based on our set-up (see Methods and Supplementary Methods for details). We compare SMINA and Vinardo scores
against the experimental binding affinities using both RMSD filter types for the two scoring functions and cutoffs of
101 Angstroms, 100 Angstroms and 3 Angstroms (\textbf{Figure 2}). The sizes of the RefinedSet{\textbackslash}CoreSet
data satisfying the filters are provided in \textbf{Supplementary Table S1}. 
\bigskip

The results indicate that the outputs of the docking tools consistently have a moderate correlation with the
experimental results (on the order of 0.5), for both the RefinedSet{\textbackslash}CoreSet, and the CoreSet. Upon
restricting the Experimental RMSD cutoff, the Pearson correlation with the experimental docking scores (both SMINA and
Vinardo) improves to 0.58-0.59 for the RefinedSet{\textbackslash}CoreSet at RMSD {\textless} 100 Angstroms, while the
correlation improves for the CoreSet only for RMSD {\textless} 3 Angstroms. The latter effect could be a result of
either a legitimate increase in prediction quality when restricting the Experimental RMSD for the CoreSet, or could
simply be the stochastic impact of a reduction in the dataset size. To test this, we ran Monte Carlo simulations
randomly subsetting 197 out of 266 complexes a thousand times, and calculating the resulting correlation with
experimental affinities. The empirical null distributions demonstrate that the improvement is unlikely to be a
statistical artifact of subsampling (\textbf{Supplementary Figure S2B}). Overall, this indicates that constraining the
pose to be similar to the experimental pose has benefits for the quality of the docking score prediction. The results
for the Consensus RMSD do not suggest any significant improvement by lowering the RMSD cutoff (only a slight
improvement is observed for the RefinedSet{\textbackslash}CoreSet).
\bigskip

Additionally, it is interesting to note that the prediction on the CoreSet is consistently better than that of the RefinedSet{\textbackslash}CoreSet. The most obvious explanation for this is that the scoring function for Vinardo was
trained on the PDBbind 2013 CoreSet. This dataset overlaps with the PDBbind 2016 CoreSet we are using. The SMINA
scoring function was trained on the CSAR-NRC HiQ 2010 data set \textsuperscript{93}, which includes some structures
from PDBbind RefinedSet from 2007. These overlaps will result in better performance on the CoreSet. Another possibility
is that the complexes in the CoreSet are, by design, of a higher quality in terms of the removal of steric clashes and
other structural irregularities relative to the RefinedSet{\textbackslash}CoreSet. Working with such protein-ligand
pairs may result in a better fit to the experimental binding affinities. We also note that the SMINA and Vinardo scores
are highly correlated with each other, a fact that impacts meta-modeling later, as two strongly collinear features
compete with each other when input into the same model.
\bigskip

We conclude that the overall performances of the docking results are not strong enough to make reliable predictions. The
results motivated us to explore alternative methods of binding affinity prediction or scoring, especially deep learning
tools.

\subsection{Deep learning models}
\subsubsection[Trained models using BindingDB or PDBbind{}-RefinedSet]{\textit{Trained models using BindingDB or
PDBbind-RefinedSet}}
\textcolor{black}{As our first group of DL models, we investigated either pre-trained (i.e., published) or
}\textit{\textcolor{black}{de novo}}\textcolor{black}{{}-trained models using the BindingDB (BDB) or PDBbind-RefinedSet
(PDBb) data (model class D1, D2, or D3, respectively) (}\textbf{\textcolor{black}{Table 2}}\textcolor{black}{). Based
on mean-squared-error (MSE), we selected the following top 6 out of the 12 }\textit{\textcolor{black}{de
novo}}\textcolor{black}{{}-trained models for subsequent analyses, which show comparable performance to the DeepPurpose
pre-trained models with MSE {\textless} 0.6: CNN-CNN, Daylight-AAC, Morgan-AAC, Morgan-CNN, MPNN-CNN, and
Transformer-CNN (}\textbf{\textcolor{black}{Supplementary Figure S3}}\textcolor{black}{). As the overlap between the
published and }\textit{\textcolor{black}{de novo}}\textcolor{black}{{}-trained models is only two, CNN-CNN and
MPNN-CNN, this expanded the family of BDB-trained base models to a total of 10 with competitive performances in this
study (}\textbf{\textcolor{black}{Table 2}}\textcolor{black}{). In terms of mean cross-validation (from 10x 5-fold CV),
the best performing BDB-pre-trained, BDB-de-novo-trained, and PDBb-trained models were Daylight-AAC, Daylight-CNN, and
CNN-AAC (}\textbf{\textcolor{black}{Table 2}}\textcolor{black}{). The model average Pearson correlation coefficients
were 0.468, 0.475, and 0.746 for the CoreSet, respectively (}\textbf{\textcolor{black}{Figure 3A}}\textcolor{black}{;
D1, D2, and D3). The two BDB-trained models perform worse than the docking tools (}\textbf{\textcolor{black}{Figure
2}}\textcolor{black}{). Full performance results are provided in }\textbf{\textcolor{black}{Supplementary Tables S2 and
S3}}\textcolor{black}{. We note that the PDBb-trained models performed poorly for BDB-Kd predictions with no
correlations (}\textbf{Supplementary Figure S4}\textcolor{black}{).}

\subsubsection[Fine{}-tuned models using PDBbind{}-RefinedSet]{\textit{Fine-tuned models using PDBbind-RefinedSet}}
\textcolor{black}{As our second group of models, we investigated fine-tuned models using PDBb for both BDB-pre-trained
and BDB-de-novo-trained models (D1F and D2F, respectively). In terms of mean cross-validation (from 10x 5-fold CV), the
best performing PDBb-finetuned models for BDB-pre-trained and BDB-de-novo-trained models were CNN-CNN and MPNN-AAC. The model average Pearson correlation coefficients were 0.716 and 0.749 for the CoreSet, respectively
(}\textbf{\textcolor{black}{Figure 3A}}\textcolor{black}{). Full performance results are provided in
}\textbf{\textcolor{black}{Supplementary Table S3}}\textcolor{black}{. As for BDB-Kd prediction, all PDBb-finetuned
models maintained good performances with Pearson correlations of 0.6–0.7, compared to predictions by the BDB-trained
models (}\textbf{\textcolor{black}{Supplementary Figures S4}}\textcolor{black}{ and
}\textbf{\textcolor{black}{S5}}\textcolor{black}{).}

\subsubsection[PCA of predictions]{\textit{PCA of predictions}}
\textcolor{black}{As our third or final group of DL models, we investigated PCA projections of all predictions from
those fine-tuned models for the BDB-pre-trained (D1F) or BDB-de-novo-trained (D2F) in the second model group
(6*5*10=300 D1F or D2F model instances from 10x 5-fold CV for each model architecture) or those PDBb-trained models
(D3) in the first model group (10*5*10=500 model instances). The first principal component (PC1) shows the best
performance in all PCA, Pearson correlation coefficients for D1F, D2F, and D3 being 0.715, 0.749, and 0.747,
respectively (}\textbf{\textcolor{black}{Figure 3A}}\textcolor{black}{; D1FP, D2FP, and D3P). We also performed PCA on
the merged set of all predictions from D1F, D2F, and D3 (300+300+500=1100 model instances), the Pearson correlation
coefficient for PC1 being 0.744 (}\textbf{\textcolor{black}{Figure 3A}}\textcolor{black}{; DAP). As expected, the PC1
is highly correlated with the mean prediction with the largest Pearson correlation coefficient of {\textgreater} 0.99
in all cases, while the second most correlated PC was not the second PC (PC2). By looking at performances of all
cross-validated fine-tuned or PCA models, we observe that PC1 and D3 models tend to perform better than D1 or D2 models
for the CoreSet in terms of Pearson correlation coefficients or mean squared errors (}\textbf{\textcolor{black}{Figure
3B}}\textcolor{black}{).}

\subsubsection[Comparisons among DL and docking models]{\textit{Comparisons among DL and docking models}}
\textcolor{black}{Given that we have investigated the 38 DL models (6 D1, 6 D2, 10 D3, 6 D1F, 6 D2F, 1 D1FP, 1 D2FP, 1
D3P, and 1 DAP) as well as the 4 docking models, we first compared their predictions by Pearson correlation values
among themselves. }\textbf{\textcolor{black}{Figure 4A}}\textcolor{black}{ shows a heatmap of Pearson correlation
values for all pairs of predictions for the CoreSet. As expected, high correlations (i.e., similar predictions) are
observed among (1) the 4 docking models and (2) the fine-tuned or PC1 models (D1F, D2F, D1FP, D2FP, D3P). The
BindingDB-based models (D1 and D2) are less correlated among themselves. Correlations between the experimental binding
affinity and those predictions show that the top model is D1F{\textbar}Daylight\_AAC (pre-trained and fine-tuned) and
the next top 3 models are PC1 models (}\textbf{\textcolor{black}{Figure 4B}}\textcolor{black}{). We also performed
another comparison analysis by prediction errors or deviations from the experimental binding affinity
(}\textbf{\textcolor{black}{Supplementary Table S3}}\textcolor{black}{). }\textbf{\textcolor{black}{Figures
4C}}\textcolor{black}{ and }\textbf{\textcolor{black}{4D}}\textcolor{black}{ show a heatmap and a box-whisker plot of
prediction deviations or absolute errors for the 42 models. The fine-tuned or PC1 models tend to show lower errors than
the BindingDB models, which in turn tend to show lower errors than the docking models. The top 5 models in terms of
median errors are the 4 PC1 models and D2F{\textbar}MPNN\_Transformer (}\textit{\textcolor{black}{de
novo}}\textcolor{black}{{}-trained and fine-tuned). Similar but more pronounced correlations and deviations are
observed for the training set, as expected (}\textbf{\textcolor{black}{Supplementary Figure S6}}\textcolor{black}{).}

\subsection[Meta{}-models of docking and DL tools]{Meta-models of docking and DL tools}
We ran several versions of meta-models to cover the spectrum of input features discussed in Methods (\textbf{Table 1}),
with the goal of parsing the contributions of particular components of the models. The broad feature groups included
docking scores, untransformed DL scores, and PC-transformed DL scores. Early exploratory analyses with
SwissADME-derived ligand properties suggested the inclusion of molecular weight to contend with possible biases in the
docking scores. Furthermore, we examined the impact of different DL training sets and models.
\bigskip

We present representative meta-model results using XGBoost in \textbf{Table 3} and \textbf{Fig. 5}, which showed the
best performance among the 4 algorithms (\textbf{Supplementary Tables S4-S6}). The different panels show the scores for
the XGBoost meta-models trained on RefinedSet{\textbackslash}CoreSet in black and the corresponding test scores for the
CoreSet in red. The panels of \textbf{Fig. 5} are arranged in a manner that reflects the different sets of features and
training data: starting with meta-models constructed entirely based on the docking methods, SMINA and Vinardo; adding
in MW; the inclusion of DL scores based on BindingDB or PDBbind trainings; and the fine-tuning and PCA. Note that, in
the PC-based models, the optimal number of PCs was obtained by maximizing the respective machine-learning scores for a
leave-out validation set (the RefinedSet{\textbackslash}CoreSet was split 80\%-20\% into a training-validation
partition).
\bigskip

We first note that the docking results alone (\textit{E}) produce a moderate Pearson correlation for the test set
(0.527). Unexpectedly, the \textit{E} correlation for the training set (0.487) is lower than that for the test set.
This might be due to the more careful curation and filtering of complexes in the CoreSet, relative to the broader
RefinedSet{\textbackslash}CoreSet. Moreover, adding in the molecular weight of the ligands (\textit{EW}) showed
significant improvements for both training and test sets. The original motivation for including the MW was the fact
that the absolute deviation between the SMINA and Vinardo predictions and the experimental binding affinity is
relatively highly correlated (r \~{} 0.3) with the MW and a few related ADME features (such as the number of heavy
atoms and molecular refractivity) than others (\textbf{Supplementary Table S7}). Including MW was an attempt at
addressing this bias in the docking scores. We also noticed that the MW was correlated with the absolute deviation in
the scores for some of the deep learning features, including the top principal component. When we added molecular
weight as a meta-model feature (in all the meta-models except for \textit{E}), we found that the deviations of the
final predictions are much less correlated with molecular weight (\textbf{Supplementary Table S8}).
\bigskip

In the next rows, it can be seen that the meta-models including DL scores significantly increase the Pearson correlation
relative to the docking-based meta-models. The fine-tuning of the BindingDB-trained models on the
PDBbind-RefinedSet{\textbackslash}CoreSet improves the prediction performance. PCA on any of the DL scores improves
performance even more. We note here that \textit{D3} was trained on the RefinedSet{\textbackslash}CoreSet, while the
\textit{D1F} and \textit{D2F} models were fine-tuned on the same set. This means that the training set for the DL
training/fine-tuning process is identical to those for the associated meta-models (\textit{ED1-F}, \textit{ED1-F-P},
\textit{ED2-F}, \textit{ED2-F-P}, \textit{ED3}, \textit{ED3-P}, and \textit{ED-A-P}). This also means that the
leave-out validation split for the PC number optimization is not an independent dataset. The consequence of the most
overlap between the DL and meta-model training sets is that we observe a very high correlation in the training set for
those meta-models. Nonetheless, the high performances of these models on the external validation (CoreSet) signal good
overall model robustness. The only caveat to this conclusion would stem from any systematic bias of the PDBbind dataset
as a whole: if all the complexes in the full dataset were biased, then the use of a common training set would enable
the models to learn the bias in the CoreSet as well, artificially inflating the performance. We are, however, not aware
of any such bias in the PDBbind dataset, given that the complexes are sourced from a large number of independent
studies. 
\bigskip

Beneficial effects observed in the meta-models by performing PCA on the DL outputs are possibly due to the denoising
or debiasing of outputs. The denoising is achieved by first running multiple parallel cross-validations, and then
combining them into a lower-dimensional representation. However, unlike the DL results in \textbf{Fig. 3}, the
performance improvement in the meta-models is not observed with the first PC alone, but rather requires the combination
of a few PCs. We also note that, among the 4 PCA-based models (\textit{ED1-F-P}, \textit{ED2-F-P}, \textit{ED3-P}, and
\textit{ED-A-P}), there are no large differences in the performance, i.e., different meta-model algorithms for PCA on
different DL training processes show similar performances (R = 0.76 - 0.78). This suggests that we have maximized the
information content of the constituent docking and DL model predictions.
\bigskip

To understand the contributions of the component features, we extracted the feature importance scores for each
meta-model (\textbf{\textcolor{black}{Supplementary }}\textbf{Figure S7}). We find that, while the contributions of the
docking scores dominate for the \textit{EW}, \textit{ED1}, and \textit{ED2}, these contributions drop to very low
values for the other DL-containing meta-models. In all meta-models, SMINA scores contribute more than Vinardo scores.
However, given the aforementioned similarity between the SMINA and Vinardo scores, this difference in the contributions
between the two docking scores is more likely to be an artifact of the modeling process (and how it deals with nearly
collinear features). The other noteworthy aspect of the importance scores is that the performance of \textit{ED3},
fine-tuned, and PC-based meta-models is dominated by one, or at most two significant features. \textit{ED1-F},
\textit{ED2-F }and \textit{ED3} are impacted mainly by two DL scores, while the PC models are dominated by PC1. We
observe again that PC1 alone is not sufficient to improve the predictions, but requires the assistance of other PCs.
The patterns of feature importance scores, however, are not entirely consistent across the different ML meta-models.
The top features of the LASSO and Linear Regression meta-models are distinct from the XGBoost top features. The signed
feature importance scores in the linear models show less domination of single scores, especially in the PC-based
models. In particular, PC1, which is individually highly correlated with experimental binding affinities, is less
important in the predictions by the linear meta-models. Given the similar performances of the ML meta-models, it
appears that there are several paths to attaining nearly equivalent overall correlation values.
\bigskip

Given the improvement in the docking predictions by filtering out certain ligand-protein pairs (\textbf{Figs. 2
}and\textbf{ S1}), we explored the effect of the Experimental and Consensus RMSD filters with 3 and 100 Angstrom
cutoffs on the meta-models (\textbf{Supplementary Tables S4 and S5}). We found that the \textit{E} and \textit{EW}
meta-models are affected by the different filters, with the Experimental RMSD filters producing slightly better results
than the Consensus filters. This reinforces the idea that the docking affinity predictions benefit from constraining
the predictions based on the pose. However, as the influence of the docking predictions decreases in the meta-models,
it appears as if the impact of the RMSD filters also goes down and the best-performing meta-models show negligible
difference in their performance with the various RMSD filters.

\subsection{Prediction synergy and complementarity of docking and DL tools}
\textcolor{black}{As we obtained better overall performance by meta-modeling of docking and DL tools, we further
investigated the degree of prediction synergy} \textcolor{black}{and complementarity of the two groups of tools, }i.e.,
whether the meta-models utilize the strengths of both tool groups and\textcolor{black}{ the degree to which the two
p}rovide non-overlapping information\textcolor{black}{. To do this, we focused on absolute errors of the mean
predictions for the benchmark set by the 4 meta-models with ED-A-P, by the 4 docking tools, and by the DAP{\textbar}PC1
model (}\textbf{\textcolor{black}{Fig. 4D}}\textcolor{black}{). When mutually compared, the mean predictions by the
meta-models achieved the lowest errors for 123 of the 266 complexes (46.2\%), the DAP{\textbar}PC1 for 104 complexes
(39.1\%), and the docking tools for 39 complexes (14.7\%) (}\textbf{\textcolor{black}{Figs. 6A-6C}}\textcolor{black}{).
If we only compare the docking and DAP{\textbar}PC1 tools, the former showed better mean predictions for 48 of the 266
complexes (18.0\%) than the latter (}\textbf{\textcolor{black}{Fig. 6D}}\textcolor{black}{). This suggests that, beyond
the globally better performance of the meta-models, prediction synergy is realized for 46.2\% of the benchmark set by
our meta-modeling due to complementary information of the docking and DL tools (even though each group of tools is
better at predicting a subset of complexes by itself). Similarly, although the DL tools are generally better than the
docking tools, there exists a small subset of complexes that are better predicted by the latter.}
\bigskip

To further parse the differences between the outputs of the prediction tools, we assessed whether certain properties of
the proteins or ligands are over- or under-represented in the complexes which were: (1) Best modeled by empirical
docking; (2) Best modeled by deep learning; (3) Best modeled by meta-modeling. We also explored the top 50 complexes as
predicted by the XGBoost Meta-model for ED-A-P to see if the performance of the meta-models relative to the
experimental data was contingent on protein or ligand properties. We extracted protein features from the UniProt
database and ligand features from the SwissADME database. We tested the significance of the occurrence of certain
annotations (using Fisher$\text{\textgreek{’}}$s exact test) and of the distributions of certain quantitative features
(using a two-sided Wilcoxon test) for each of the four subsets of complexes. Specifically, we considered whether
several UniProt features were annotated more or less frequently, and whether certain Gene Ontology (GO)
\textsuperscript{94}, domain and pathway annotations occurred with different probabilities in the subsets. We also
compared the distributions of the ligand SwissADME features. The results (\textbf{Supplementary Table S9}) indicate
that there are practically no significant (nominal p-value  ${\leq}0.05$) protein features prioritized in the four
cases considered. This may be due to the fact that several proteins (and their associated UniProt IDs) are duplicated
in the CoreSet. The number of reviewed proteins with UniProtIDs drops from 243 to 60 after removal of all duplicates.
The highly significant results with nominal p-values  ${\leq}0.01$\ (both Wilcoxon and t-tests) are the distributions
of several ligand features including MW, lipophilicity, and solubility measures in the structures for which the
empirical docking tools perform the best. There is also some hint of significance for MW, the “\#Rotatable bonds” and
“\#H-bond acceptors” in the structures for which the meta-models perform the best.

\subsection[Comparison with structure{}-based tools]{Comparison with structure-based tools}
\textcolor{black}{Having built our meta-models based on sequence-based DL models, we were interested in comparisons with
structure-based tools through voxelization of binding pockets. As described in
}\textbf{\textcolor{black}{Methods}}\textcolor{black}{, we focused on the following 3 DL tools by re-training them with
our training data for fair comparisons as much as possible as well as one recent tool based on advanced mathematics.
HAC-Net }\textcolor{black}{\textsuperscript{47}}\textcolor{black}{ was trained on the PDBbind-2020 general set of
{\textgreater}18,800 ligand-protein complexes. Their best performance was PCC = 0.846 for the
PDBbind-2016 CoreSet of 290 complexes. We note that we focused on the PDBbind-2020 RefinedSet rather than the general
set because of two reasons: (1) higher quality and (2) limited resources. The re-trained HAC-Net with our training set
showed PCC = 0.756 for the CoreSet we used, which is worse than our best meta-model performance of PCC = 0.777 or our average DL performance of PCC = 0.763 (\textbf{Table 4}). Although their MSE value is better than ours, this 
performance metric is not a fair metric for comparison in this study because they used -log10(Kd/Ki) (also known as pKd/Ki) while we used ln(Kd/Ki), which show different distributions of binding affinities for model training. We note that the HAC-Net performance 
based on ln(Kd/Ki) notably degraded to PCC = 0.152 and RMSE = 18.85. FAST}\textcolor{black}{\textsuperscript{42}}\textcolor{black}{ was trained on the PDBbind-2016 general set excluding the CoreSet. We tested the pre-trained SG-CNN of FAST against the CoreSet we used, which resulted in PCC = 0.716, worse than HAC-Net (\textbf{Table4}). KDeep }\textcolor{black}{\textsuperscript{46}}\textcolor{black}{ was one of the earlier structure-based 3D-CNN models. Their reported Pearson correlation on the PDBbind-2016 CoreSet was 0.82, which we were not able to verify because their code is not open-source. With 3 re-trained KDeep models using 3 limited training sets from KdeepTrainer (see }\textbf{\textcolor{black}{Methods}}\textcolor{black}{), We obtained PCC = 
0.718, 0.730, and 0.754 (mean = 0.734) for the CoreSet we used (\textbf{Table 4}), which are similar to those reported values of 0.701–0.738 by Kwon et al. using the PDBbind-2016 RefinedSet as a training set and 4 different learning rates
}\textcolor{black}{\textsuperscript{50}}\textcolor{black}{. }PerSpect ML \textsuperscript{69} recently reported the
highest Pearson correlation of 0.84 in the case of the PDBbind 2016 CoreSet. While we were not able to confirm their
result due to incomplete code availability, those recent methods based on advanced mathematics from the Xia group
\textsuperscript{67-70} showed largely better performances than ours and others. We note, however, that they may
generate more than 100,000 features requiring 3D structures or binding poses, which are often unavailable in real-world
drug discovery.\textcolor{black}{ Despite these limited comparisons, we conclude that our meta-models and our
sequence-based DL models show competitive performances or advantages to more sophisticated structure-based models.}

\subsection{GeneralSet benchmark}
In addition to the CASF-2016 benchmark, we tested our meta-models against the PBDbind2020 GeneralSet of 5,261 complexes
as a benchmark (\textbf{Methods}; \textbf{Table 4}). Although the size of the GeneralSet benchmark is as large as our
training data, we observe performance improvements of our meta-models over the individual base models
(\textbf{Supplementary Figure S8}). As expected, the performance for this benchmark is worse than that for CASF-2016
(PCC {\textless}0.60 vs. {\textgreater}0.76) because of its lower data quality and the data size being similar to the
training data. As for the structure-based tools mentioned above, Pearson$\text{\textgreek{’}}$s correlation by the
re-trained HAC-Net was R = 0.008 (excluding 22 complexes that failed in pre-processing) and that by KDeep (the web-server
default model of ACEBind2023-GraphNet as of Sept. 2, 2024) was R = 0.318 (excluding 71 complexes that failed in
pre-processing), which performed worse than our meta-models or DL models and similar to the docking tools. We note that
the huge RMSE by HAC-Net is due to poor performances of the 3D-CNN component of HAC-Net. A similarly poor performance was 
observed by the published pre-trained HAC-Net with the PDBbind2020 general set (R = 0.009 and RMSE = 1138.51). The PDBbind2016-GeneralSet pre-trained SG-CNN of FAST achieved Pearson$\text{\textgreek{’}}$s correlation of 0.526 (excluding 124 complexes that failed in pre-processing), which is also a worse performance than our meta-models and DL models. Although their RMSE is better than ours, it is not a fair metric for comparison as explained in the previous section. Also, we note that 3,314 out of 5,261 complexes in our benchmark set are part of the training set of PDBbind2016 GeneralSet by FAST.

\subsection{Ranking benchmark}
To further explore the performance of the models on specific target proteins with multiple ligands, we designed a
binding affinity ranking benchmark against 3 target proteins for 35 complexes from the PBDbind v2020 GeneralSet:
O60341, O15151, and P0C6U8 (see \textbf{Methods}). The performances of our models for the 3 target proteins are shown
in \textbf{Table 5} in comparison with the 3 tools described above, HAC-Net (re-trained), SG-CNN (FAST; pre-trained),
and KDeep (the up-to-date default version on their web server). The pre-trained SG-CNN performed best for O60341 and
O15151 in terms of RMSE (2.10 and 1.37, respectively). This was expected because their training data were the PDBbind
v2016 GeneralSet, which includes 30 out of the 35 complexes in our benchmark set except 5 complexes for O60341.
Otherwise, our meta-models or fine-tuned DL models performed best for all 3 target proteins in terms of both RMSE and
Pearson$\text{\textgreek{’}}$s correlation. Our best performing models are ED2-F\_VvS\_101.0\_LinReg (meta-model;
docking + MW + de novo-trained \& fined-tuned DL by linear regression) and D1F{\textbar}Daylight\_AAC (pre-trained \&
fine-tuned DL model) for O60341 (RMSE = 2.34), EW\_VvS\_101.0\_LinReg (meta-model; docking + MW by linear regression)
for O15151 (RMSE = 1.44), and D2F{\textbar}MPNN-AAC (de novo-trained \& fine-tuned DL model) for P0C6U8 (RMSE = 1.01).
We consider the Pearson correlation metric less reliable due to the small sample size. 
\bigskip

As noted in the previous section on prediction synergy, different meta-models or component DL/docking models may perform
best for different complexes or proteins. In the case of this virtual screening, we noticed that while some of the same
meta-models would often predict the binding affinities for O60341 and P0C6U8 with reasonable accuracy, they would not
predict the affinities for O15151 as well, and vice versa. For example, ED2-F\_VvS\_101.0\_LinReg showed low RMSEs for
O60341 (RMSE = 2.34) and P0C6U8 (RMSE = 2.54) but higher RMSEs for O15151 (RMSE = 5.88). In particular, we observe that
6 out of the 8 ligands in the complexes for O15151 were k-mers (k = 10, 12, or 15) with relatively high MW of
{\textgreater}1000. Among the 35 complexes there are 18 k-mers with mean MW of 927.2 compared to 762.7 for 17
non-k-mers. We note that about 5\% of the RefinedSet (we used for model training) and about 10\% of the GeneralSet
contain k-mers (up to 10-mer and 19-mer, respectively). Given that k-mers are not representative and tend to have high
MW, we aimed to assess whether high or low MW causes differences in the performance of the meta-models independently
from the choice of a target protein by dividing ligands into two groups of low MW {\textless}= 900 and high MW
{\textgreater} 900 (the median threshold for the 35 complexes). By focusing on ED2-F\_VvS\_101.0\_LinReg from above, we
find that the accuracy for low MW ligands is better than high MW ligands (RMSE = 2.45 vs. 4.36). This is expected given
the fact that the ligands in the RefinedSet we used for training have mean MW of 384.3 and standard deviation of 172.6
and that the Pearson correlations between predictions and MW are negative ranging from -0.62 to -0.29. While far from
conclusive for generalization, these results suggest that there might be identifiable features of ligands and/or
proteins which would explain why certain meta-models perform variably for different target proteins.

\subsection{Application to virtual screening}
Although the above results suggest that top-ranking high-affinity ligands predicted by our models could be prioritized
as hits in virtual screening, we also tested our models using the virtual screening benchmark dataset of LIT-PCBA to
distinguish active ligands from inactive ligands for 15 proteins (see \textbf{Methods}; \textbf{Supplementary Table
S10}). As our models were not trained for binary classification of active and inactive ligands, our hypothesis was that
active ligands tend to have higher affinity scores than inactive ligands. To test the hypothesis, we evaluated our
models and the 2 docking tools using 3 performance metrics (top5 recall, top10 recall, and precision) for each protein
(\textbf{Methods}; \textbf{Table 6}) and the distribution difference of predicted affinity scores between all active
and all inactive ligands for each protein (Welch$\text{\textgreek{’}}$s t-test and the Mann-Whitney U test;
\textbf{Supplementary Table S11}). The 3 performance metrics show that our averaged DL models and meta-models tend to
be better at predicting active ligands than the 2 docking tools (i.e., {\textgreater}50\% in blue in \textbf{Table 6}).
The docking tools performed better for 3 proteins, but their performance values are mostly 20\% or less otherwise (in
red in \textbf{Table 6}). As expected, active ligands tend to have higher affinity scores than inactive ligands for
most of the proteins individually, and when aggregated across all proteins (\textbf{Supplementary Table S11}). We
emphasize that, as in the section on prediction synergy, there is variability among the target proteins in terms of the
best-performing model. Different protein targets may benefit from the use of different modeling techniques, conditional
on their physicochemical properties. This is also observed in terms of the target-specific behavior of the p-values
(\textbf{Supplementary Table S11}). All those structure-based models mentioned above, HAC-Net, FAST, KDeep, and the
models from the Xia group (e.g. PerSpect ML), require input data of docked poses and hence cannot be used in the
LIT-PCBA virtual screening benchmark or are not generally applicable to identification of active vs. inactive ligands.
\bigskip

Overall, we conclude that the family of our models are competitive to several state-of-the-art structure-based models
and also useful for virtual screening applications.

\clearpage\section{Discussion}
The framework we present here, inspired by stacked generalization and super learner \textsuperscript{74, 75}, is based
on the central concept that different models designed and trained for the same prediction task often carry
model-specific biases. Hence, simple approaches to combining the results of their independent predictions may yield
superior results if the biases are even partially independent and the models are complementary to some degree. Indeed,
we find this to be borne out in the ligand-protein binding affinity prediction in several ways. First, we find that
combining the predictions from empirical docking methods with those from the pre-trained or \textit{de novo}{}-trained
deep learning tools can yield gains in predictive power over the individual approaches (ED1-3 vs. E or D1-2). Second,
utilizing models trained on the BindingDB as the base and refining them on the PDBbind resulted in significant
improvements (D1 and D2 vs. D1F and D2F, respectively), demonstrating that fine-tuning pre-trained models could retain
good predictive quality on both primary (BindingDB) and secondary (PDBbind) datasets
(\textbf{\textcolor{black}{Supplementary }}\textbf{Fig. S4}). Third, even within the suite of deep learning tools,
transforming their prediction scores through principal component analysis could improve the performance on the test set
(i.e., principal component regression). This may be due to a de-noising of individual DL scores. Finally, in addition
to performance improvement or synergy by the meta-models over the docking and DL models for a subset of complexes, the
use of meta-models by canonical machine learning enables us to explicitly add features that may be associated with
prediction deviations from the experimental affinities in the individual models. For example, we found that, for both
empirical docking tools and some deep learning models, the deviations of the predictions from the experimental
affinities were correlated with molecular weight. Explicitly adding in molecular weight as a meta-model feature
resulted in the deviations of the final predictions being much less correlated with molecular weight. Moreover, our DL
models and meta-models demonstrated their competitive power in comparison to other recently developed structure-based
models in a large-scale affinity prediction benchmark and a virtual screening application benchmark.
\bigskip

Overall, our framework corroborates existing work on ensemble learning, stacked generalization, super learner, and
consensus-based improvements in predictions. Our machine-learning-based meta-models allow for the flexibility of adding
or removing any component tools, physicochemical properties, or molecular descriptors as part of feature engineering.
Models trained on multiple datasets (e.g. BindingDB and PDBbind) and on multiple data types (e.g. sequence and/or
structure) may be unified to reduce individual biases and improve predictions. Furthermore, we posit that the DL models
from DeepPurpose considered herein, being sequence-based, offer better scalability. Given that only ligand and protein
sequences are required for the prediction, and not all-atom 3D structures, we expect that increasingly large databases
could be used for training in future iterations by running high-throughput experimental binding affinity assays
(without the need for concurrent X-ray or NMR experiments) as a filtering strategy in early drug discovery. Considering
that the observed performances of these models for all our benchmarks are better or on par with more sophisticated
structure-based DL models, there is an argument to be made in favor of more scalable approaches. Additionally, we note
that the public and private experimental binding affinity databases are highly fragmented, in that they are produced
separately, with potentially different types of assays, or at different scales. While efforts such as the PDBbind
database manage to carefully curate a high-quality single dataset, not all data sources can be incorporated because of
the type of data (e.g. experimental data without associated structures being available) or because of data access (e.g.
behind a private paywall). Our meta-modeling may address some of these issues by pooling only prediction outputs from
models trained on diverse public or private databases with or without available structures, protecting the databases
themselves if necessary and maintaining performance quality. Subsequent in-depth analysis could also identify
prediction synergy and complementarity of all available tools as we demonstrated in \textbf{Fig. 6}. Finally, there
might be value in expanding families of meta-models with different input data, each family being optimized to target a
particular protein or class of proteins and/or ligands as demonstrated in \textbf{Tables 5 and 6}.
\bigskip

We acknowledge some limitations of our study. First, we primarily focused on a single dataset, PDBbind RefinedSet, for
model training. In fact, in light of recent positive results from training models on the full PDBbind GeneralSet of
\~{}19,000 complexes \textsuperscript{42, 47}, there is a scope for further scale-up of the model training process with
larger data. It is worth mentioning, however, that when Jones et al. used the PDBbind general or refined set for model
training, they observed the tradeoffs between data quality and quantity in performance of different models: SG-CNN
performed better with the general set (in favor of data quantity); 3D-CNN performed better with the refined set (in
favor of data quality). On the other hand, we started with DL models pre-trained on BindingDB of \~{}66,000 complexes
and either employed them “out of the box” or fine-tuned them on the PDBbind RefinedSet (excluding the CoreSet). The two
BindingDB-based meta-models, ED1 and ED2, perform better than any of the stand-alone docking or DL tools (Pearson
correlation 0.62-0.64 vs. 0.47-0.59). We also tested the two PDBbind fine-tuned models, D1F and D2F, on the primary
training set of BindingDB, to ascertain the degree to which the models retained their “out of the box” performance on
the primary training set. We found reasonably good correlations for D1F and D2F (r = 0.636 to 0.711)
(\textbf{\textcolor{black}{Supplementary }}\textbf{Fig. S4}) compared to the pre-trained D1 (r = 0.8)
\textsuperscript{57}. However, the PDBbind-trained D3 showed poor prediction on BindingDB (r {\textless} 0.2). More
importantly, while the CASF-2016 benchmark set has been well accepted, more benchmark sets of high quality would be
needed to test and compare models in a more objective manner and also to challenge model generalization. Our second
benchmark using the large-scale GeneralSet, although lower quality than CASF-2016, was such an effort to assess model
generalization. Our virtual screening benchmark is also very limited, although our models outperformed the docking
tools and tend to perform well for ligands with low molecular weight. In particular, the binary classification problem
for screening of active or functional ligands would require a new model development for higher accuracy rather than
repurposing exclusively affinity-based regression models as in this study. In this sense, the good performances of our
fine-tuned DL models or meta-models do not imply model generalizability in an absolute sense. We also note that we
neither fine-tuned the docking tools nor conducted hyperparameter optimization for the DL models, which might help
performance as well.
\bigskip

Our proof-of-concept study of meta-modeling can be readily extended to integrate with (i.e., to add more meta-features
of) predictions from any other tools, such as HAC-Net, FAST, KDeep, and/or PerSpect ML, molecular properties or
embeddings, or even predictions from the same base models trained on different orthogonal datasets, to address model
bias and robustness, which is the key conceptual advantage of any ensemble or fusion model. In parallel to this type of
horizontal extension (i.e., feature engineering), it is conceivable that meta-models may be expanded or stacked
vertically, leading to deep meta-models. Our meta-modeling approach offers promising and flexible strategies for future
model iterations to further improve ligand-protein binding affinity prediction.

\clearpage\section{Data and code availability}
The BindingDB and PDBbind databases are publicly available. The code is partially available, due to our pending patent,
on our GitHub page at \url{https://github.com/Lee1701/Lee2023a}.
\bigskip

\section{Supporting Information}
Supporting Information contains supplementary methods, supplementary figures, and supplementary references.
\bigskip

\section{Acknowledgements}
We thank the Yale Center for Research Computing for their high-performance-computing resources and help. We also thank
Drs. Jeffrey Brock and Sarah Miller for their support in the initiation of this project through the COVID HASTE
community at the Yale School of Engineering \& Applied Science.
\bigskip

\section{Author contributions}
HL and PSE conceived, designed, and performed the study, analyzed the data, and drafted the manuscript. HL, PSE, and MGB
interpreted the data and critically reviewed and approved the manuscript. HL supervised the study.
\bigskip

\section{Competing interests}
HL reports a consulting role at Guidepoint outside of this submitted work. The other authors report no competing
interests.

\clearpage\section{References}
1.\ \ Shen, C.; \ Ding, J.; \ Wang, Z.; \ Cao, D.; \ Ding, X.; Hou, T., From machine learning to deep learning: Advances
in scoring functions for protein–ligand docking. \textit{WIREs Computational Molecular Science }\textbf{2020,}
\textit{10}, e1429.

2.\ \ Ain, Q. U.; \ Aleksandrova, A.; \ Roessler, F. D.; Ballester, P. J., Machine-learning scoring functions to improve
structure-based binding affinity prediction and virtual screening. \textit{Wiley Interdiscip Rev Comput Mol Sci
}\textbf{2015,} \textit{5}, 405-424.

3.\ \ Khamis, M. A.; \ Gomaa, W.; Ahmed, W. F., Machine learning in computational docking. \textit{Artif Intell Med
}\textbf{2015,} \textit{63}, 135-52.

4.\ \ Kitchen, D. B.; \ Decornez, H.; \ Furr, J. R.; Bajorath, J., Docking and scoring in virtual screening for drug
discovery: methods and applications. \textit{Nat Rev Drug Discov }\textbf{2004,} \textit{3}, 935-49.

5.\ \ Moitessier, N.; \ Englebienne, P.; \ Lee, D.; \ Lawandi, J.; Corbeil, C. R., Towards the development of universal,
fast and highly accurate docking/scoring methods: a long way to go. \textit{Br J Pharmacol }\textbf{2008,} \textit{153
Suppl 1}, S7-26.

6.\ \ Grinter, S. Z.; Zou, X., Challenges, applications, and recent advances of protein-ligand docking in
structure-based drug design. \textit{Molecules }\textbf{2014,} \textit{19}, 10150-76.

7.\ \ Su, M.; \ Yang, Q.; \ Du, Y.; \ Feng, G.; \ Liu, Z.; \ Li, Y.; Wang, R., Comparative Assessment of Scoring
Functions: The CASF-2016 Update. \textit{J Chem Inf Model }\textbf{2019,} \textit{59}, 895-913.

8.\ \ Ferreira, L. G.; \ Dos Santos, R. N.; \ Oliva, G.; Andricopulo, A. D., Molecular docking and structure-based drug
design strategies. \textit{Molecules }\textbf{2015,} \textit{20}, 13384-421.

9.\ \ Sousa, S. F.; \ Fernandes, P. A.; Ramos, M. J., Protein-ligand docking: current status and future challenges.
\textit{Proteins }\textbf{2006,} \textit{65}, 15-26.

10.\ \ Krovat, M. E.; \ Steindl, T.; Langer, T., Recent Advances in Docking and Scoring. \textit{Current Computer-Aided
Drug Design }\textbf{2005,} \textit{1}, 93-102.

11.\ \ Ballester, P. J.; Mitchell, J. B. O., A machine learning approach to predicting protein–ligand binding affinity
with applications to molecular docking. \textit{Bioinformatics }\textbf{2010,} \textit{26}, 1169-1175.

12.\ \ Böhm, H. J., The development of a simple empirical scoring function to estimate the binding constant for a
protein-ligand complex of known three-dimensional structure. \textit{J Comput Aided Mol Des }\textbf{1994,} \textit{8},
243-56.

13.\ \ Ruvinsky, A. M., Role of binding entropy in the refinement of protein-ligand docking predictions: analysis based
on the use of 11 scoring functions. \textit{J Comput Chem }\textbf{2007,} \textit{28}, 1364-72.

14.\ \ Mysinger, M. M.; Shoichet, B. K., Rapid context-dependent ligand desolvation in molecular docking. \textit{J Chem
Inf Model }\textbf{2010,} \textit{50}, 1561-73.

15.\ \ Lill, M. A., Efficient incorporation of protein flexibility and dynamics into molecular docking simulations.
\textit{Biochemistry }\textbf{2011,} \textit{50}, 6157-69.

16.\ \ Cheng, T.; \ Li, X.; \ Li, Y.; \ Liu, Z.; Wang, R., Comparative Assessment of Scoring Functions on a Diverse Test
Set. \textit{Journal of Chemical Information and Modeling }\textbf{2009,} \textit{49}, 1079-1093.

17.\ \ Wang, R.; \ Lu, Y.; \ Fang, X.; Wang, S., An Extensive Test of 14 Scoring Functions Using the PDBbind Refined Set
of 800 Protein$-$Ligand Complexes. \textit{Journal of Chemical Information and Computer Sciences }\textbf{2004,}
\textit{44}, 2114-2125.

18.\ \ Warren, G. L.; \ Andrews, C. W.; \ Capelli, A. M.; \ Clarke, B.; \ LaLonde, J.; \ Lambert, M. H.; \ Lindvall, M.;
\ Nevins, N.; \ Semus, S. F.; \ Senger, S.; \ Tedesco, G.; \ Wall, I. D.; \ Woolven, J. M.; \ Peishoff, C. E.; Head, M.
S., A critical assessment of docking programs and scoring functions. \textit{J Med Chem }\textbf{2006,} \textit{49},
5912-31.

19.\ \ Trott, O.; Olson, A. J., AutoDock Vina: improving the speed and accuracy of docking with a new scoring function,
efficient optimization, and multithreading. \textit{J Comput Chem }\textbf{2010,} \textit{31}, 455-61.

20.\ \ Koes, D. R.; \ Baumgartner, M. P.; Camacho, C. J., Lessons learned in empirical scoring with smina from the CSAR
2011 benchmarking exercise. \textit{J Chem Inf Model }\textbf{2013,} \textit{53}, 1893-904.

21.\ \ Wang, R.; Wang, S., How does consensus scoring work for virtual library screening? An idealized computer
experiment. \textit{J Chem Inf Comput Sci }\textbf{2001,} \textit{41}, 1422-6.

22.\ \ Charifson, P. S.; \ Corkery, J. J.; \ Murcko, M. A.; Walters, W. P., Consensus scoring: A method for obtaining
improved hit rates from docking databases of three-dimensional structures into proteins. \textit{J Med Chem
}\textbf{1999,} \textit{42}, 5100-9.

23.\ \ Houston, D. R.; Walkinshaw, M. D., Consensus Docking: Improving the Reliability of Docking in a Virtual Screening
Context. \textit{Journal of Chemical Information and Modeling }\textbf{2013,} \textit{53}, 384-390.

24.\ \ Palacio-Rodríguez, K.; \ Lans, I.; \ Cavasotto, C. N.; Cossio, P., Exponential consensus ranking improves the
outcome in docking and receptor ensemble docking. \textit{Scientific Reports }\textbf{2019,} \textit{9}, 5142.

25.\ \ Liu, Z.; \ Su, M.; \ Han, L.; \ Liu, J.; \ Yang, Q.; \ Li, Y.; Wang, R., Forging the Basis for Developing
Protein–Ligand Interaction Scoring Functions. \textit{Accounts of Chemical Research }\textbf{2017,} \textit{50},
302-309.

26.\ \ Ashtawy, H. M.; Mahapatra, N. R., A Comparative Assessment of Ranking Accuracies of Conventional and
Machine-Learning-Based Scoring Functions for Protein-Ligand Binding Affinity Prediction. \textit{IEEE/ACM Trans.
Comput. Biol. Bioinformatics }\textbf{2012,} \textit{9}, 1301–1313.

27.\ \ Patel, L.; \ Shukla, T.; \ Huang, X.; \ Ussery, D. W.; Wang, S., Machine Learning Methods in Drug Discovery.
\textit{Molecules }\textbf{2020,} \textit{25}.

28.\ \ Wójcikowski, M.; \ Ballester, P. J.; Siedlecki, P., Performance of machine-learning scoring functions in
structure-based virtual screening. \textit{Sci Rep }\textbf{2017,} \textit{7}, 46710.

29.\ \ Baum, B.; \ Muley, L.; \ Smolinski, M.; \ Heine, A.; \ Hangauer, D.; Klebe, G., Non-additivity of functional
group contributions in protein-ligand binding: a comprehensive study by crystallography and isothermal titration
calorimetry. \textit{J Mol Biol }\textbf{2010,} \textit{397}, 1042-54.

30.\ \ Baskin, II; \ Winkler, D.; Tetko, I. V., A renaissance of neural networks in drug discovery. \textit{Expert Opin
Drug Discov }\textbf{2016,} \textit{11}, 785-95.

31.\ \ Dana, D.; \ Gadhiya, S. V.; \ St Surin, L. G.; \ Li, D.; \ Naaz, F.; \ Ali, Q.; \ Paka, L.; \ Yamin, M. A.;
\ Narayan, M.; \ Goldberg, I. D.; Narayan, P., Deep Learning in Drug Discovery and Medicine; Scratching the Surface.
\textit{Molecules }\textbf{2018,} \textit{23}.

32.\ \ Sapoval, N.; \ Aghazadeh, A.; \ Nute, M. G.; \ Antunes, D. A.; \ Balaji, A.; \ Baraniuk, R.; \ Barberan, C. J.;
\ Dannenfelser, R.; \ Dun, C.; \ Edrisi, M.; \ Elworth, R. A. L.; \ Kille, B.; \ Kyrillidis, A.; \ Nakhleh, L.;
\ Wolfe, C. R.; \ Yan, Z.; \ Yao, V.; Treangen, T. J., Current progress and open challenges for applying deep learning
across the biosciences. \textit{Nature Communications }\textbf{2022,} \textit{13}, 1728.

33.\ \ Yang, Z.; \ Zeng, X.; \ Zhao, Y.; Chen, R., AlphaFold2 and its applications in the fields of biology and
medicine. \textit{Signal Transduction and Targeted Therapy }\textbf{2023,} \textit{8}, 115.

34.\ \ Akdel, M.; \ Pires, D. E. V.; \ Pardo, E. P.; \ Jänes, J.; \ Zalevsky, A. O.; \ Mészáros, B.; \ Bryant, P.;
\ Good, L. L.; \ Laskowski, R. A.; \ Pozzati, G.; \ Shenoy, A.; \ Zhu, W.; \ Kundrotas, P.; \ Serra, V. R.;
\ Rodrigues, C. H. M.; \ Dunham, A. S.; \ Burke, D.; \ Borkakoti, N.; \ Velankar, S.; \ Frost, A.; \ Basquin, J.;
\ Lindorff-Larsen, K.; \ Bateman, A.; \ Kajava, A. V.; \ Valencia, A.; \ Ovchinnikov, S.; \ Durairaj, J.; \ Ascher, D.
B.; \ Thornton, J. M.; \ Davey, N. E.; \ Stein, A.; \ Elofsson, A.; \ Croll, T. I.; Beltrao, P., A structural biology
community assessment of AlphaFold2 applications. \textit{Nature Structural \& Molecular Biology }\textbf{2022,}
\textit{29}, 1056-1067.

35.\ \ Abbasi, K.; \ Razzaghi, P.; \ Poso, A.; \ Ghanbari-Ara, S.; Masoudi-Nejad, A., Deep Learning in Drug Target
Interaction Prediction: Current and Future Perspectives. \textit{Curr Med Chem }\textbf{2021,} \textit{28}, 2100-2113.

36.\ \ Li, H.; \ Zou, L.; \ Kowah, J. A. H.; \ He, D.; \ Liu, Z.; \ Ding, X.; \ Wen, H.; \ Wang, L.; \ Yuan, M.; Liu,
X., A compact review of progress and prospects of deep learning in drug discovery. \textit{J Mol Model }\textbf{2023,}
\textit{29}, 117.

37.\ \ Kim, J.; \ Park, S.; \ Min, D.; Kim, W., Comprehensive Survey of Recent Drug Discovery Using Deep Learning.
\textit{Int J Mol Sci }\textbf{2021,} \textit{22}.

38.\ \ Lavecchia, A., Deep learning in drug discovery: opportunities, challenges and future prospects. \textit{Drug
Discov Today }\textbf{2019,} \textit{24}, 2017-2032.

39.\ \ Ma, J.; \ Sheridan, R. P.; \ Liaw, A.; \ Dahl, G. E.; Svetnik, V., Deep Neural Nets as a Method for Quantitative
Structure–Activity Relationships. \textit{Journal of Chemical Information and Modeling }\textbf{2015,} \textit{55},
263-274.

40.\ \ Krizhevsky, A.; \ Sutskever, I.; Hinton, G. E., ImageNet classification with deep convolutional neural networks.
In \textit{Proceedings of the 25th International Conference on Neural Information Processing Systems - Volume 1},
Curran Associates Inc.: Lake Tahoe, Nevada, 2012; pp 1097–1105.

41.\ \ Szegedy, C.; \ Liu, W.; \ Jia, Y.; \ Sermanet, P.; \ Reed, S.; \ Anguelov, D.; \ Erhan, D.; \ Vanhoucke, V.;
Rabinovich, A. Going Deeper with Convolutions 2014, p. arXiv:1409.4842.
\url{https://ui.adsabs.harvard.edu/abs/2014arXiv1409.4842S} (accessed September 01, 2014).

42.\ \ Jones, D.; \ Kim, H.; \ Zhang, X.; \ Zemla, A.; \ Stevenson, G.; \ Bennett, W. F. D.; \ Kirshner, D.; \ Wong, S.
E.; \ Lightstone, F. C.; Allen, J. E., Improved Protein–Ligand Binding Affinity Prediction with Structure-Based Deep
Fusion Inference. \textit{Journal of Chemical Information and Modeling }\textbf{2021,} \textit{61}, 1583-1592.

43.\ \ Gomes, J.; \ Ramsundar, B.; \ Feinberg, E. N.; Pande, V. S. Atomic Convolutional Networks for Predicting
Protein-Ligand Binding Affinity 2017, p. arXiv:1703.10603. \url{https://ui.adsabs.harvard.edu/abs/2017arXiv170310603G}
(accessed March 01, 2017).

44.\ \ Wallach, I.; \ Dzamba, M.; Heifets, A. AtomNet: A Deep Convolutional Neural Network for Bioactivity Prediction in
Structure-based Drug Discovery 2015, p. arXiv:1510.02855. \url{https://ui.adsabs.harvard.edu/abs/2015arXiv151002855W}
(accessed October 01, 2015).

45.\ \ Ragoza, M.; \ Hochuli, J.; \ Idrobo, E.; \ Sunseri, J.; Koes, D. R., Protein–Ligand Scoring with Convolutional
Neural Networks. \textit{Journal of Chemical Information and Modeling }\textbf{2017,} \textit{57}, 942-957.

46.\ \ Jiménez, J.; \ Škalič, M.; \ Martínez-Rosell, G.; De Fabritiis, G., KDEEP: Protein–Ligand Absolute Binding
Affinity Prediction via 3D-Convolutional Neural Networks. \textit{Journal of Chemical Information and Modeling
}\textbf{2018,} \textit{58}, 287-296.

47.\ \ Kyro, G. W.; \ Brent, R. I.; Batista, V. S., HAC-Net: A Hybrid Attention-Based Convolutional Neural Network for
Highly Accurate Protein–Ligand Binding Affinity Prediction. \textit{Journal of Chemical Information and Modeling
}\textbf{2023,} \textit{63}, 1947-1960.

48.\ \ Wang, Z.; \ Zheng, L.; \ Liu, Y.; \ Qu, Y.; \ Li, Y.-Q.; \ Zhao, M.; \ Mu\leavevmode\allowbreak\,, Y.;
Li\leavevmode\allowbreak\,, W., OnionNet-2: A Convolutional Neural Network Model for Predicting Protein-Ligand Binding
Affinity Based on Residue-Atom Contacting Shells. \textit{Frontiers in Chemistry }\textbf{2021,} \textit{9}.

49.\ \ Shan, W.; \ Li, X.; \ Yao, H.; Lin, K., Convolutional Neural Network-based Virtual Screening. \textit{Curr Med
Chem }\textbf{2021,} \textit{28}, 2033-2047.

50.\ \ Kwon, Y.; \ Shin, W.-H.; \ Ko, J.; Lee, J., AK-Score: Accurate Protein-Ligand Binding Affinity Prediction Using
an Ensemble of 3D-Convolutional Neural Networks. \textit{International Journal of Molecular Sciences }\textbf{2020,}
\textit{21}, 8424.

51.\ \ Jiang, H.; \ Wang, J.; \ Cong, W.; \ Huang, Y.; \ Ramezani, M.; \ Sarma, A.; \ Dokholyan, N. V.; \ Mahdavi, M.;
Kandemir, M. T., Predicting Protein-Ligand Docking Structure with Graph Neural Network. \textit{J Chem Inf Model
}\textbf{2022,} \textit{62}, 2923-2932.

52.\ \ Knutson, C.; \ Bontha, M.; \ Bilbrey, J. A.; Kumar, N., Decoding the protein-ligand interactions using parallel
graph neural networks. \textit{Sci Rep }\textbf{2022,} \textit{12}, 7624.

53.\ \ Nikolaienko, T.; \ Gurbych, O.; Druchok, M., Complex machine learning model needs complex testing: Examining
predictability of molecular binding affinity by a graph neural network. \textit{J Comput Chem }\textbf{2022,}
\textit{43}, 728-739.

54.\ \ Yang, Z.; \ Zhong, W.; \ Lv, Q.; \ Dong, T.; Yu-Chian Chen, C., Geometric Interaction Graph Neural Network for
Predicting Protein-Ligand Binding Affinities from 3D Structures (GIGN). \textit{J Phys Chem Lett }\textbf{2023,}
\textit{14}, 2020-2033.

55.\ \ Feinberg, E. N.; \ Sur, D.; \ Wu, Z.; \ Husic, B. E.; \ Mai, H.; \ Li, Y.; \ Sun, S.; \ Yang, J.; \ Ramsundar,
B.; Pande, V. S., PotentialNet for Molecular Property Prediction. \textit{ACS Cent Sci }\textbf{2018,} \textit{4},
1520-1530.

56.\ \ Karlov, D. S.; \ Sosnin, S.; \ Fedorov, M. V.; Popov, P., graphDelta: MPNN Scoring Function for the Affinity
Prediction of Protein-Ligand Complexes. \textit{ACS Omega }\textbf{2020,} \textit{5}, 5150-5159.

57.\ \ Huang, K.; \ Fu, T.; \ Glass, L.; \ Zitnik, M.; \ Xiao, C.; Sun, J., DeepPurpose: a Deep Learning Library for
Drug-Target Interaction Prediction and Applications to Repurposing and Screening. \textit{arXiv }\textbf{2020,}
\textit{2004.08919}.

58.\ \ Öztürk, H.; \ Özgür, A.; Ozkirimli, E., DeepDTA: deep drug–target binding affinity prediction.
\textit{Bioinformatics }\textbf{2018,} \textit{34}, i821-i829.

59.\ \ Castelvecchi, D., Can we open the black box of AI? \textit{Nature }\textbf{2016,} \textit{538}, 20-23.

60.\ \ Doshi-Velez, F.; Kim, B., Towards a rigorous science of interpretable machine learning. \textit{arXiv preprint
arXiv:1702.08608 }\textbf{2017}.

61.\ \ Murdoch, W. J.; \ Singh, C.; \ Kumbier, K.; \ Abbasi-Asl, R.; Yu, B., Definitions, methods, and applications in
interpretable machine learning. \textit{Proceedings of the National Academy of Sciences }\textbf{2019,} \textit{116},
22071-22080.

62.\ \ Barredo Arrieta, A.; \ Díaz-Rodríguez, N.; \ Del Ser, J.; \ Bennetot, A.; \ Tabik, S.; \ Barbado, A.; \ Garcia,
S.; \ Gil-Lopez, S.; \ Molina, D.; \ Benjamins, R.; \ Chatila, R.; Herrera, F., Explainable Artificial Intelligence
(XAI): Concepts, taxonomies, opportunities and challenges toward responsible AI. \textit{Information Fusion
}\textbf{2020,} \textit{58}, 82-115.

63.\ \ Roscher, R.; \ Bohn, B.; \ Duarte, M. F.; Garcke, J., Explainable Machine Learning for Scientific Insights and
Discoveries. \textit{IEEE Access }\textbf{2020,} \textit{8}, 42200-42216.

64.\ \ Cang, Z.; Wei, G. W., Integration of element specific persistent homology and machine learning for protein-ligand
binding affinity prediction. \textit{Int J Numer Method Biomed Eng }\textbf{2018,} \textit{34}.

65.\ \ Cang, Z.; \ Mu, L.; Wei, G. W., Representability of algebraic topology for biomolecules in machine learning based
scoring and virtual screening. \textit{PLoS Comput Biol }\textbf{2018,} \textit{14}, e1005929.

66.\ \ Wu, J.; \ Chen, H.; \ Cheng, M.; Xiong, H., CurvAGN: Curvature-based Adaptive Graph Neural Networks for
Predicting Protein-Ligand Binding Affinity. \textit{BMC Bioinformatics }\textbf{2023,} \textit{24}, 378.

67.\ \ Wee, J.; Xia, K., Ollivier Persistent Ricci Curvature-Based Machine Learning for the Protein–Ligand Binding
Affinity Prediction. \textit{Journal of Chemical Information and Modeling }\textbf{2021,} \textit{61}, 1617-1626.

68.\ \ Wee, J.; Xia, K., Forman persistent Ricci curvature (FPRC)-based machine learning models for protein–ligand
binding affinity prediction. \textit{Briefings in Bioinformatics }\textbf{2021,} \textit{22}.

69.\ \ Meng, Z.; Xia, K., Persistent spectral–based machine learning (PerSpect ML) for protein-ligand binding affinity
prediction. \textit{Science Advances }\textbf{2021,} \textit{7}, eabc5329.

70.\ \ Liu, X.; \ Feng, H.; \ Wu, J.; Xia, K., Persistent spectral hypergraph based machine learning (PSH-ML) for
protein-ligand binding affinity prediction. \textit{Briefings in Bioinformatics }\textbf{2021,} \textit{22}.

71.\ \ Nguyen, D. D.; \ Gao, K.; \ Wang, M.; Wei, G.-W., MathDL: mathematical deep learning for D3R Grand Challenge 4.
\textit{Journal of Computer-Aided Molecular Design }\textbf{2020,} \textit{34}, 131-147.

72.\ \ Moon, S.; \ Zhung, W.; \ Yang, S.; \ Lim, J.; Kim, W. Y., PIGNet: a physics-informed deep learning model toward
generalized drug–target interaction predictions. \textit{Chemical Science }\textbf{2022,} \textit{13}, 3661-3673.

73.\ \ Guedes, I. A.; \ Barreto, A. M. S.; \ Marinho, D.; \ Krempser, E.; \ Kuenemann, M. A.; \ Sperandio, O.;
\ Dardenne, L. E.; Miteva, M. A., New machine learning and physics-based scoring functions for drug discovery.
\textit{Scientific Reports }\textbf{2021,} \textit{11}, 3198.

74.\ \ Wolpert, D. H., Stacked generalization. \textit{Neural Networks }\textbf{1992,} \textit{5}, 241-259.

75.\ \ Laan, M. J. v. d.; \ Polley, E. C.; Hubbard, A. E., Super Learner. \textit{Statistical Applications in Genetics
and Molecular Biology }\textbf{2007,} \textit{6}.

76.\ \ Bennett, J.; \ Elkan, C.; \ Liu, B.; \ Smyth, P.; Tikk, D., KDD Cup and workshop 2007. \textit{SIGKDD Explor.
Newsl. }\textbf{2007,} \textit{9}, 51–52.

77.\ \ Feuerverger, A.; \ He, Y.; Khatri, S., Statistical Significance of the Netflix Challenge. \textit{Statistical
Science }\textbf{2012,} \textit{27}, 202-231, 30.

78.\ \ Hallinan, B.; Striphas, T., Recommended for you: The Netflix Prize and the production of algorithmic culture.
\textit{New Media \& Society }\textbf{2016,} \textit{18}, 117-137.

79.\ \ Bell, R. M.; \ Koren, Y.; Volinsky, C., All Together Now: A Perspective on the Netflix Prize. \textit{CHANCE
}\textbf{2010,} \textit{23}, 24-29.

80.\ \ Reichstein, M.; \ Camps-Valls, G.; \ Stevens, B.; \ Jung, M.; \ Denzler, J.; \ Carvalhais, N.; Prabhat, Deep
learning and process understanding for data-driven Earth system science. \textit{Nature }\textbf{2019,} \textit{566},
195-204.

81.\ \ Karniadakis, G. E.; \ Kevrekidis, I. G.; \ Lu, L.; \ Perdikaris, P.; \ Wang, S.; Yang, L., Physics-informed
machine learning. \textit{Nature Reviews Physics }\textbf{2021,} \textit{3}, 422-440.

82.\ \ Thuerey, N.; \ Holl, P.; \ Mueller, M.; \ Schnell, P.; \ Trost, F.; Um, K., Physics-based Deep Learning.
\textit{ArXiv }\textbf{2021,} \textit{abs/2109.05237}.

83.\ \ Wang, B.; \ Yan, C.; \ Lou, S.; \ Emani, P.; \ Li, B.; \ Xu, M.; \ Kong, X.; \ Meyerson, W.; \ Yang, Y. T.;
\ Lee, D.; Gerstein, M., Building a Hybrid Physical-Statistical Classifier for Predicting the Effect of Variants
Related to Protein-Drug Interactions. \textit{Structure }\textbf{2019,} \textit{27}, 1469-1481.e3.

84.\ \ Gilson, M. K.; \ Liu, T.; \ Baitaluk, M.; \ Nicola, G.; \ Hwang, L.; Chong, J., BindingDB in 2015: A public
database for medicinal chemistry, computational chemistry and systems pharmacology. \textit{Nucleic Acids Research
}\textbf{2015,} \textit{44}, D1045-D1053.

85.\ \ Tran-Nguyen, V.-K.; \ Jacquemard, C.; Rognan, D., LIT-PCBA: An Unbiased Data Set for Machine Learning and Virtual
Screening. \textit{Journal of Chemical Information and Modeling }\textbf{2020,} \textit{60}, 4263-4273.

86.\ \ Li, Y.; \ Han, L.; \ Liu, Z.; Wang, R., Comparative Assessment of Scoring Functions on an Updated Benchmark: 2.
Evaluation Methods and General Results. \textit{Journal of Chemical Information and Modeling }\textbf{2014,}
\textit{54}, 1717-1736.

87.\ \ Quiroga, R.; Villarreal, M. A., Vinardo: A Scoring Function Based on Autodock Vina Improves Scoring, Docking, and
Virtual Screening. \textit{PLoS One }\textbf{2016,} \textit{11}, e0155183.

88.\ \ Blanes-Mira, C.; \ Fernández-Aguado, P.; \ de Andrés-López, J.; \ Fernández-Carvajal, A.; \ Ferrer-Montiel, A.;
Fernández-Ballester, G., Comprehensive Survey of Consensus Docking for High-Throughput Virtual Screening.
\textit{Molecules }\textbf{2023,} \textit{28}, 175.

89.\ \ Daina, A.; \ Michielin, O.; Zoete, V., SwissADME: a free web tool to evaluate pharmacokinetics, drug-likeness and
medicinal chemistry friendliness of small molecules. \textit{Scientific Reports }\textbf{2017,} \textit{7}, 42717.

90.\ \ The UniProt Consortium, UniProt: the Universal Protein Knowledgebase in 2023. \textit{Nucleic Acids Res
}\textbf{2023,} \textit{51}, D523-d531.

91.\ \ Torrens-Fontanals, M.; \ Tourlas, P.; \ Doerr, S.; De Fabritiis, G., PlayMolecule Viewer: A Toolkit for the
Visualization of Molecules and Other Data. \textit{Journal of Chemical Information and Modeling }\textbf{2024,}
\textit{64} (3), 584-589.

92.\ \ Camacho, C.; \ Coulouris, G.; \ Avagyan, V.; \ Ma, N.; \ Papadopoulos, J.; \ Bealer, K.; Madden, T. L., BLAST+:
architecture and applications. \textit{BMC Bioinformatics }\textbf{2009,} \textit{10}, 421.

93.\ \ Dunbar, J. B., Jr.; \ Smith, R. D.; \ Yang, C. Y.; \ Ung, P. M.; \ Lexa, K. W.; \ Khazanov, N. A.; \ Stuckey, J.
A.; \ Wang, S.; Carlson, H. A., CSAR benchmark exercise of 2010: selection of the protein-ligand complexes. \textit{J
Chem Inf Model }\textbf{2011,} \textit{51}, 2036-46.

94.\ \ Aleksander, S. A.; \ Balhoff, J.; \ Carbon, S.; \ Cherry, J. M.; \ Drabkin, H. J.; \ Ebert, D.; \ Feuermann, M.;
\ Gaudet, P.; \ Harris, N. L.; \ Hill, D. P.; \ Lee, R.; \ Mi, H.; \ Moxon, S.; \ Mungall, C. J.; \ Muruganugan, A.;
\ Mushayahama, T.; \ Sternberg, P. W.; \ Thomas, P. D.; \ Van Auken, K.; \ Ramsey, J.; \ Siegele, D. A.; \ Chisholm, R.
L.; \ Fey, P.; \ Aspromonte, M. C.; \ Nugnes, M. V.; \ Quaglia, F.; \ Tosatto, S.; \ Giglio, M.; \ Nadendla, S.;
\ Antonazzo, G.; \ Attrill, H.; \ Dos Santos, G.; \ Marygold, S.; \ Strelets, V.; \ Tabone, C. J.; \ Thurmond, J.;
\ Zhou, P.; \ Ahmed, S. H.; \ Asanitthong, P.; \ Luna Buitrago, D.; \ Erdol, M. N.; \ Gage, M. C.; \ Ali Kadhum, M.;
\ Li, K. Y. C.; \ Long, M.; \ Michalak, A.; \ Pesala, A.; \ Pritazahra, A.; \ Saverimuttu, S. C. C.; \ Su, R.;
\ Thurlow, K. E.; \ Lovering, R. C.; \ Logie, C.; \ Oliferenko, S.; \ Blake, J.; \ Christie, K.; \ Corbani, L.;
\ Dolan, M. E.; \ Drabkin, H. J.; \ Hill, D. P.; \ Ni, L.; \ Sitnikov, D.; \ Smith, C.; \ Cuzick, A.; \ Seager, J.;
\ Cooper, L.; \ Elser, J.; \ Jaiswal, P.; \ Gupta, P.; \ Jaiswal, P.; \ Naithani, S.; \ Lera-Ramirez, M.; \ Rutherford,
K.; \ Wood, V.; \ De Pons, J. L.; \ Dwinell, M. R.; \ Hayman, G. T.; \ Kaldunski, M. L.; \ Kwitek, A. E.;
\ Laulederkind, S. J. F.; \ Tutaj, M. A.; \ Vedi, M.; \ Wang, S. J.; \ D'Eustachio, P.; \ Aimo, L.; \ Axelsen, K.;
\ Bridge, A.; \ Hyka-Nouspikel, N.; \ Morgat, A.; \ Aleksander, S. A.; \ Cherry, J. M.; \ Engel, S. R.; \ Karra, K.;
\ Miyasato, S. R.; \ Nash, R. S.; \ Skrzypek, M. S.; \ Weng, S.; \ Wong, E. D.; \ Bakker, E.; \ Berardini, T. Z.;
\ Reiser, L.; \ Auchincloss, A.; \ Axelsen, K.; \ Argoud-Puy, G.; \ Blatter, M. C.; \ Boutet, E.; \ Breuza, L.;
\ Bridge, A.; \ Casals-Casas, C.; \ Coudert, E.; \ Estreicher, A.; \ Livia Famiglietti, M.; \ Feuermann, M.; \ Gos, A.;
\ Gruaz-Gumowski, N.; \ Hulo, C.; \ Hyka-Nouspikel, N.; \ Jungo, F.; \ Le Mercier, P.; \ Lieberherr, D.; \ Masson, P.;
\ Morgat, A.; \ Pedruzzi, I.; \ Pourcel, L.; \ Poux, S.; \ Rivoire, C.; \ Sundaram, S.; \ Bateman, A.;
\ Bowler-Barnett, E.; \ Bye, A. J. H.; \ Denny, P.; \ Ignatchenko, A.; \ Ishtiaq, R.; \ Lock, A.; \ Lussi, Y.;
\ Magrane, M.; \ Martin, M. J.; \ Orchard, S.; \ Raposo, P.; \ Speretta, E.; \ Tyagi, N.; \ Warner, K.; \ Zaru, R.;
\ Diehl, A. D.; \ Lee, R.; \ Chan, J.; \ Diamantakis, S.; \ Raciti, D.; \ Zarowiecki, M.; \ Fisher, M.; \ James-Zorn,
C.; \ Ponferrada, V.; \ Zorn, A.; \ Ramachandran, S.; \ Ruzicka, L.; Westerfield, M., The Gene Ontology knowledgebase
in 2023. \textit{Genetics }\textbf{2023,} \textit{224}.


\newpage
\textbf{Table 1. Descriptions of meta-model feature groups\\}
\includegraphics[width=6.5126in,height=6.6047in]{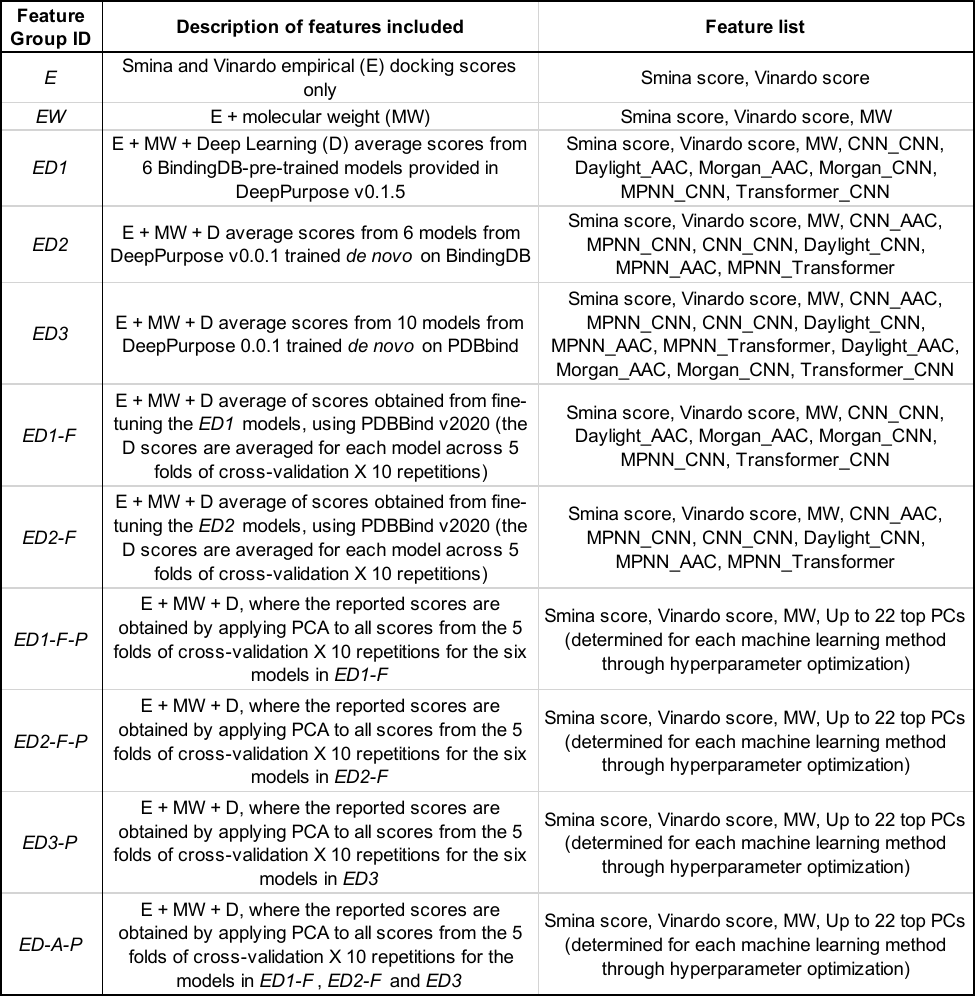}
\par

\newpage
\textbf{Table 2. Performances of DL models\\}
\includegraphics[width=6.5043in,height=7.8236in]{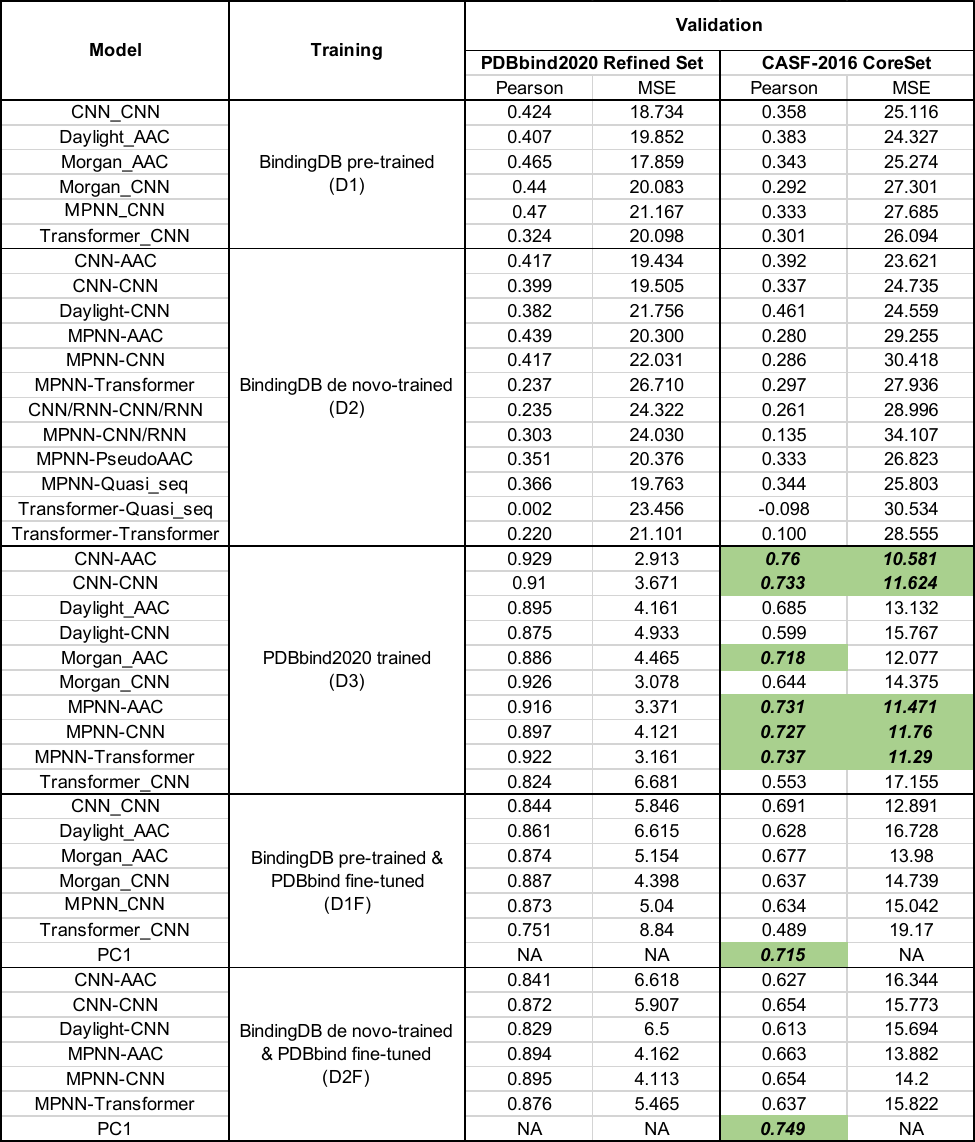}
\par
 
\newpage
\textbf{Table 3. Meta-model performances by XGBoost\\}
\includegraphics[width=6.5043in,height=2.1681in]{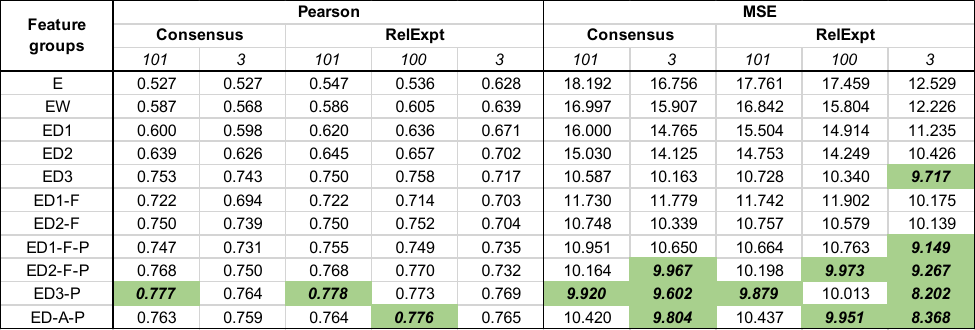}
\par
{\raggedright
\textup{N.B. Consensus = Consensus RMSD filtering; RelExpt = Experimental RMSD filtering. The values, 101, 100, and 3, represent
RMSD cutoffs in Angstroms.}
\par}

\newpage
\textbf{Table 4. Performance comparisons of a PDBbind2020 GeneralSet benchmark\\}
\includegraphics[width=6.5in,height=1.2in]{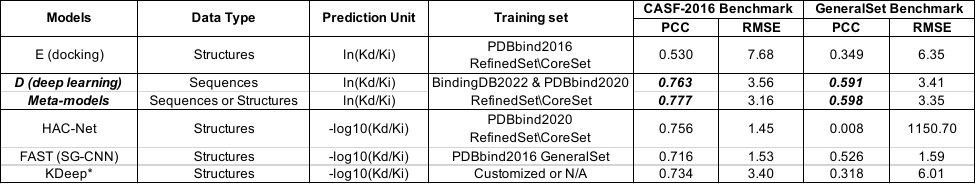}
\par
{\raggedright
\textup{N.B. PCC = Pearson correlation coefficient. RMSE = Root Mean Square Error. *Re-training for the CASF-2016 benchmark was done using 3 customized sets of PDBbind2020 RefinedSet{\textbackslash}CoreSet. The GeneralSet benchmark was based on the default KDeep model of ACEBind-2023 Graph-Net where the training set is not available (see the main text).}
\par}

\newpage
\textbf{Table 5. Performance comparisons of a ranking benchmark\\}
\includegraphics[width=6.5291in,height=1.8319in]{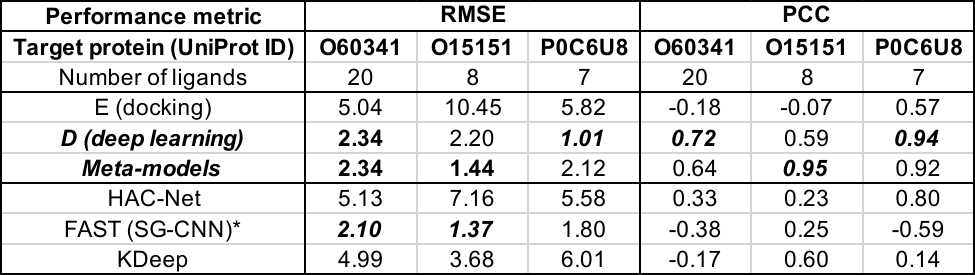}
\par
{\raggedright
\textup{N.B. The best performance value for each target protein is in bold italic, while performance values for our models are
in bold. PCC = Pearson correlation coefficient. *Pre-trained on PDBbind v2016 GeneralSet.}
\par}

\newpage
\textbf{Table 6. Enrichment of active ligands in the top predicted ligands for LIT-PCBA virtual screening\\}
\includegraphics[width=6.5043in,height=1.4035in]{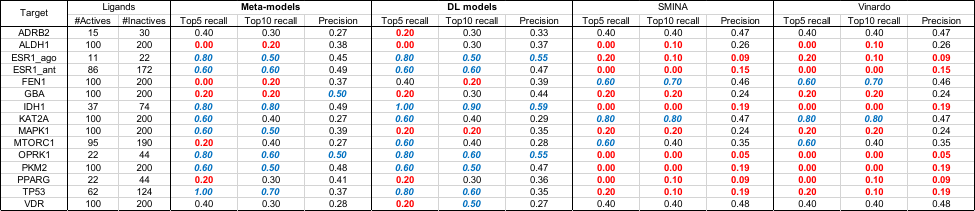}
\par
{\raggedright
\textup{N.B. {\textgreater}= 0.50 in blue; {\textless}= 0.20 in red.}
\par}


\newpage

\centering
\includegraphics[width=6.5in,height=5.3929in]{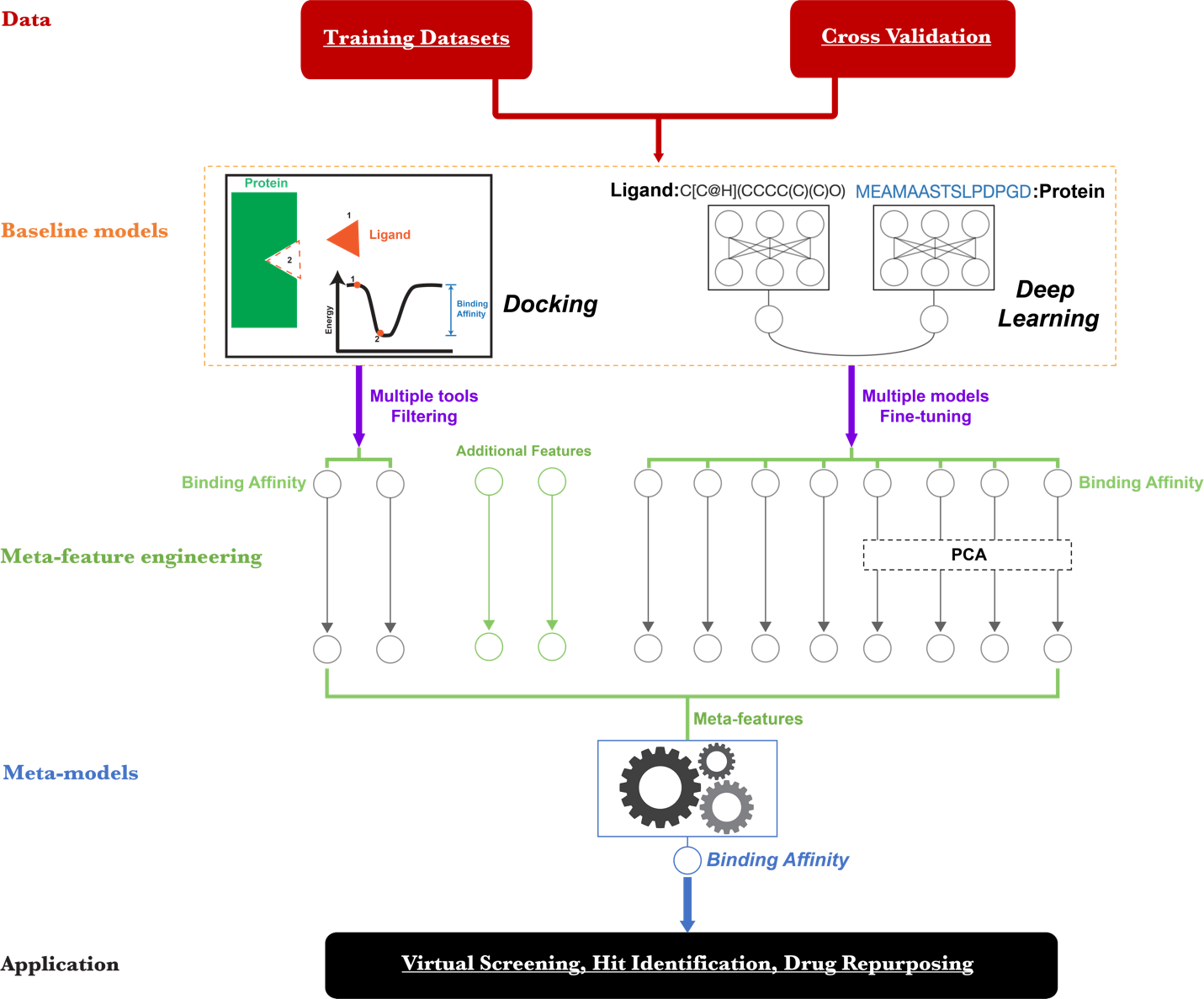} 
\par\bigskip
\raggedright\textbf{Figure 1. Overview of our workflow.} The inputs are training datasets (BindingDB and PDBbind) and a
cross-validation strategy. The baseline models consist of docking tools (SMINA and Vinardo) and deep-learning (DL)
models (from the DeepPurpose library). The DL models were either pre-trained or de novo trained, while the docking
tools were pre-trained. The output ligand poses of the docking tools are filtered by structural comparisons with the
experimental docked poses, or by comparisons between the poses predicted by the two docking tools. The
BindingDB-trained DL models are fine-tuned on the PDBbind dataset, and their prediction scores are further transformed
through principal component analysis (PCA). Additional features, such as molecular weight, can be added, before various
combinations of all these meta-features (filtered or unfiltered docking scores, fine-tuned and/or PCA-transformed DL
scores) are fed into several machine-learning meta-models for final training on the PDBbind dataset. The meta-models
include linear regression, LASSO, ElasticNet, and XGBoost.

\clearpage\centering
\includegraphics[width=6.5in,height=3.2244in]{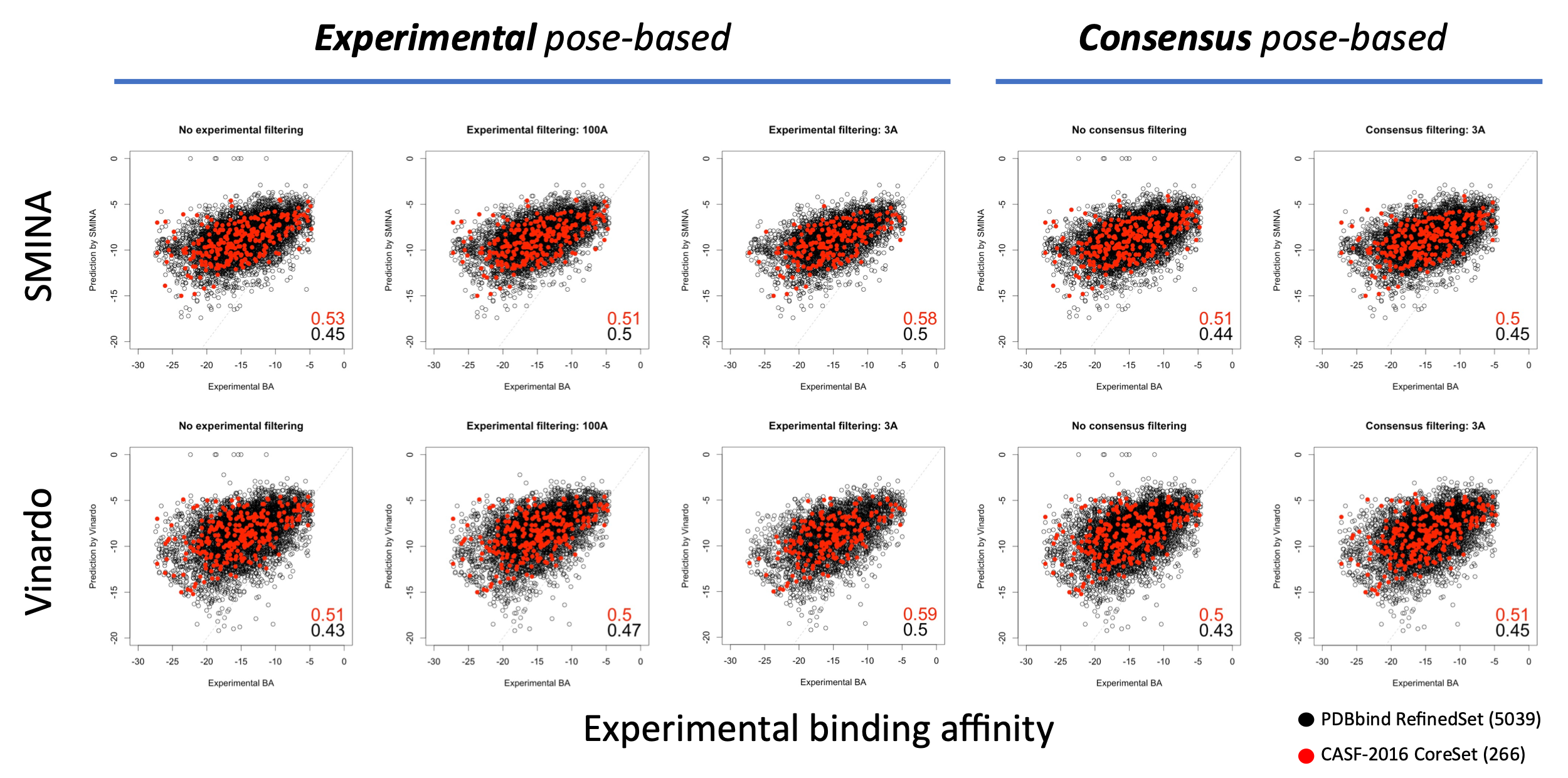} 
\par\bigskip
\raggedright\textbf{Figure 2. Performances of empirical scoring functions and consensus filtering.} The outputs of the docking
tools, SMINA and Vinardo, are shown with the binding affinities for the RefinedSet{\textbackslash}CoreSet in black and
those for the CoreSet in red. The columns indicate the different choices of filtering approaches and pose-based RMSD
thresholds. Pearson$\text{\textgreek{’}}$s correlation values for the two different sets of predictions are shown in
each panel in their respective colors.

\clearpage\centering
\lfbox[margin-bottom=0.0043in,margin-top=0mm,margin-right=0mm,margin-left=0mm,border-style=none,padding=0mm]{\includegraphics[width=4.298in,height=6.6075in]{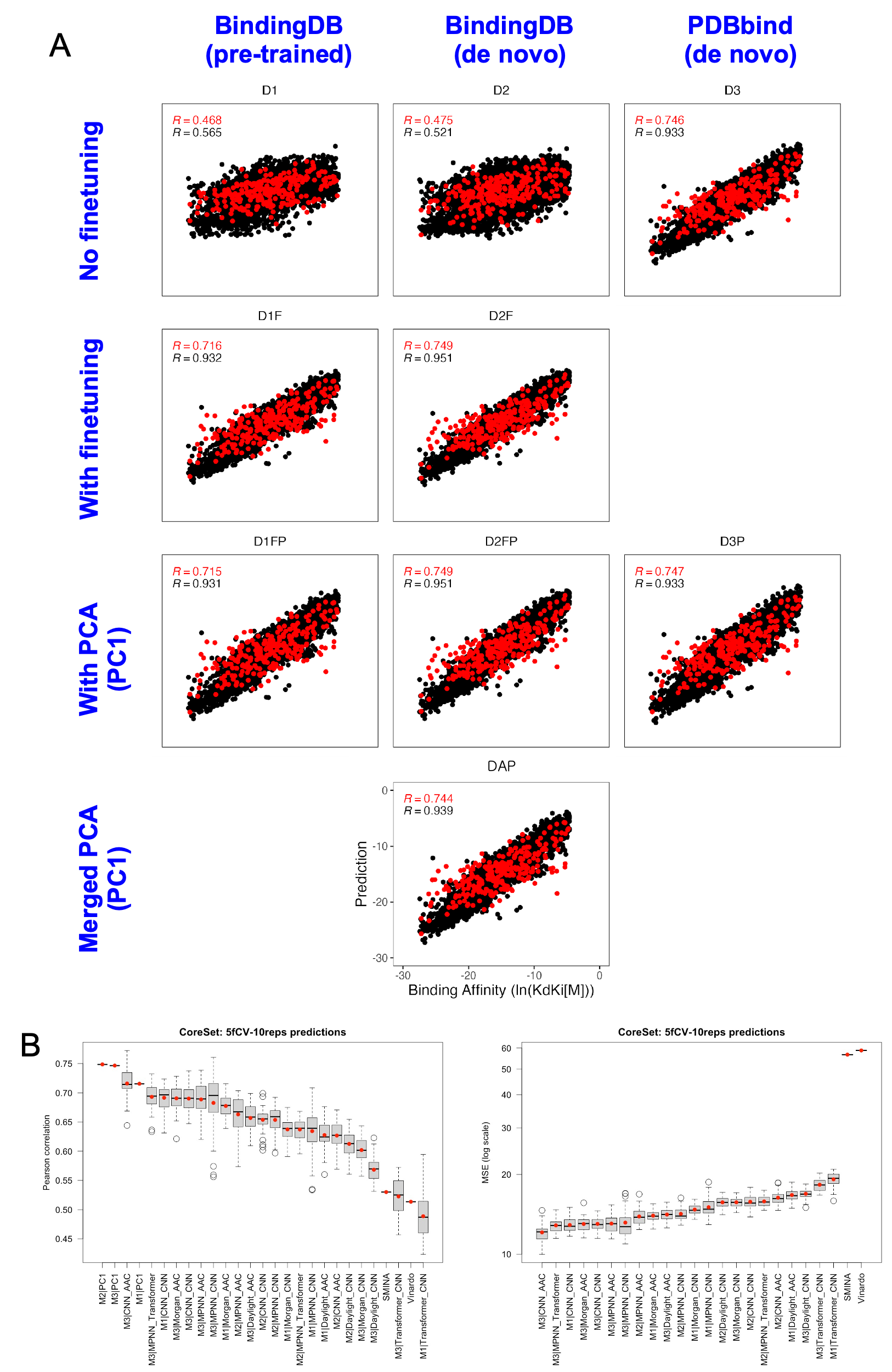}}
\par\bigskip
\raggedright\textbf{Figure 3. Performances of deep learning models.} (A) Scatter plots of experimental binding affinities (x axis)
and model predictions (y axis) for the full 5,305 complexes in the PDBbind RefinedSet (266 complexes in the CoreSet
benchmark). Each plot shows predictions by the average of the best models or PC1 in terms of Pearson correlation
coefficients for the CoreSet in each training strategy. (B) Box-whisker plots of Pearson correlation coefficients (left
panel) and mean squared errors (MSE; right panel) for the CoreSet by a total of 22 PDBbind-finetuned (M1 and M2) and
PDBbind-trained (M3) models and PC1 models of M1, M2, and M3 along with SMINA and Vinardo models. Predictions of the 22
models are mean values from 10 repeated 5-fold CVs. M1, M2, and M3 correspond to D1F, D2F, and D3 in (A).

\clearpage\centering
\lfbox[margin-bottom=0.0055in,margin-top=0mm,margin-right=0mm,margin-left=0mm,border-style=none,padding=0mm]{\includegraphics[width=6.3in,height=6.6in]{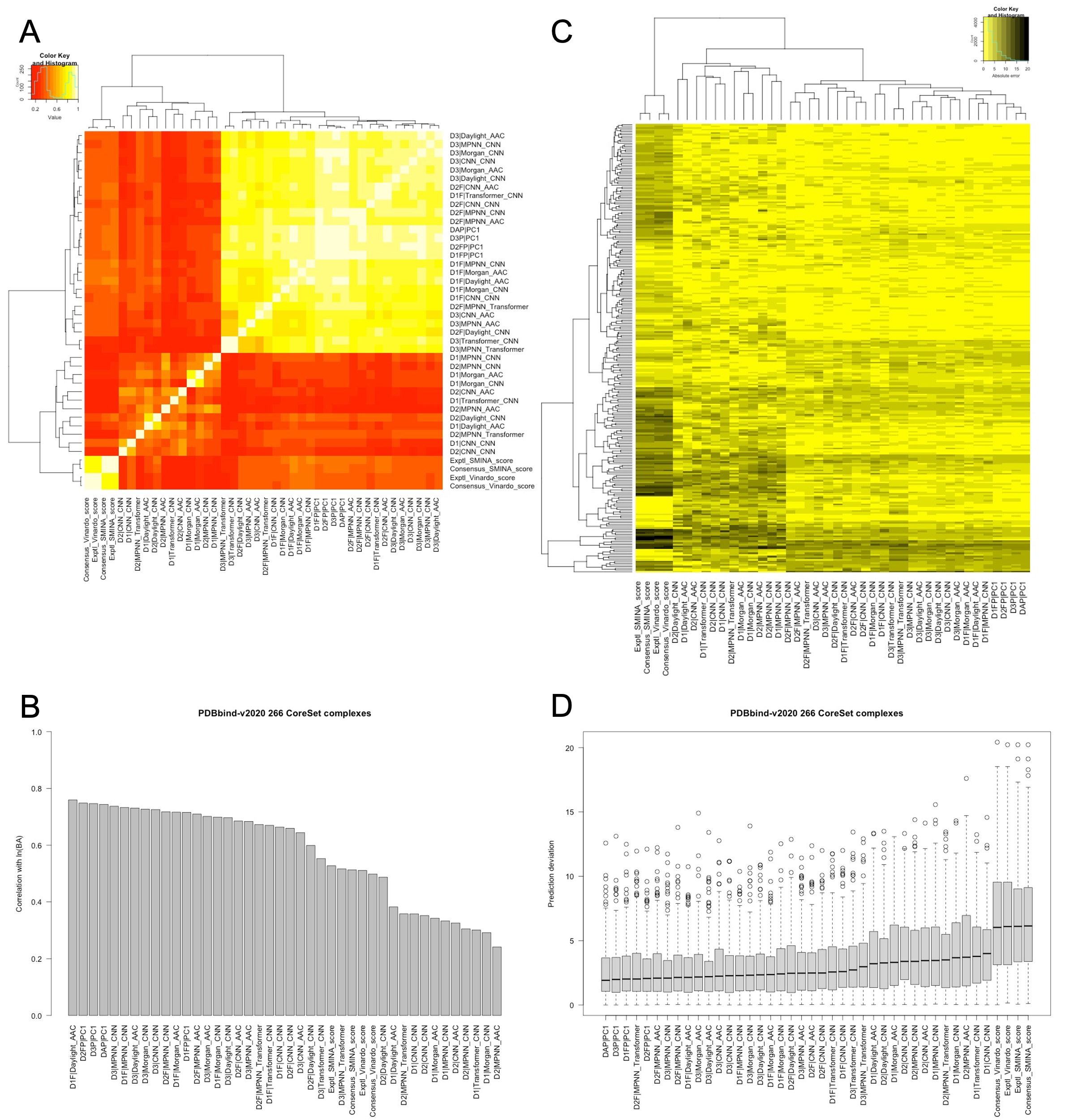}}
\par\bigskip
\raggedright\textbf{Figure 4. Model comparison of CoreSet predictions.} (A) Heatmap of Pearson correlations between all pairs of
CoreSet predictions by the 42 models from the docking and deep learning tools. The dendrograms are results of
hierarchical clustering with Euclidean distance and the complete linkage method. (B) Bar plot of Pearson correlations
between the experimental binding affinity and predictions by the 42 models for the CoreSet complexes. The models are
ordered by the correlation values. (C) Heatmap of absolute errors or deviations between the experimental binding
affinity and predictions by the 42 models (columns) for each of the CoreSet complexes (rows). The dendrograms are as in
(A). (D) Box-whisker plot of the data used in (B). The models are ordered by the median values.

\clearpage\centering
\lfbox[margin-bottom=0.0035in,margin-top=0mm,margin-right=0mm,margin-left=0mm,border-style=none,padding=0mm]{\includegraphics[width=5.0in,height=7.5in]{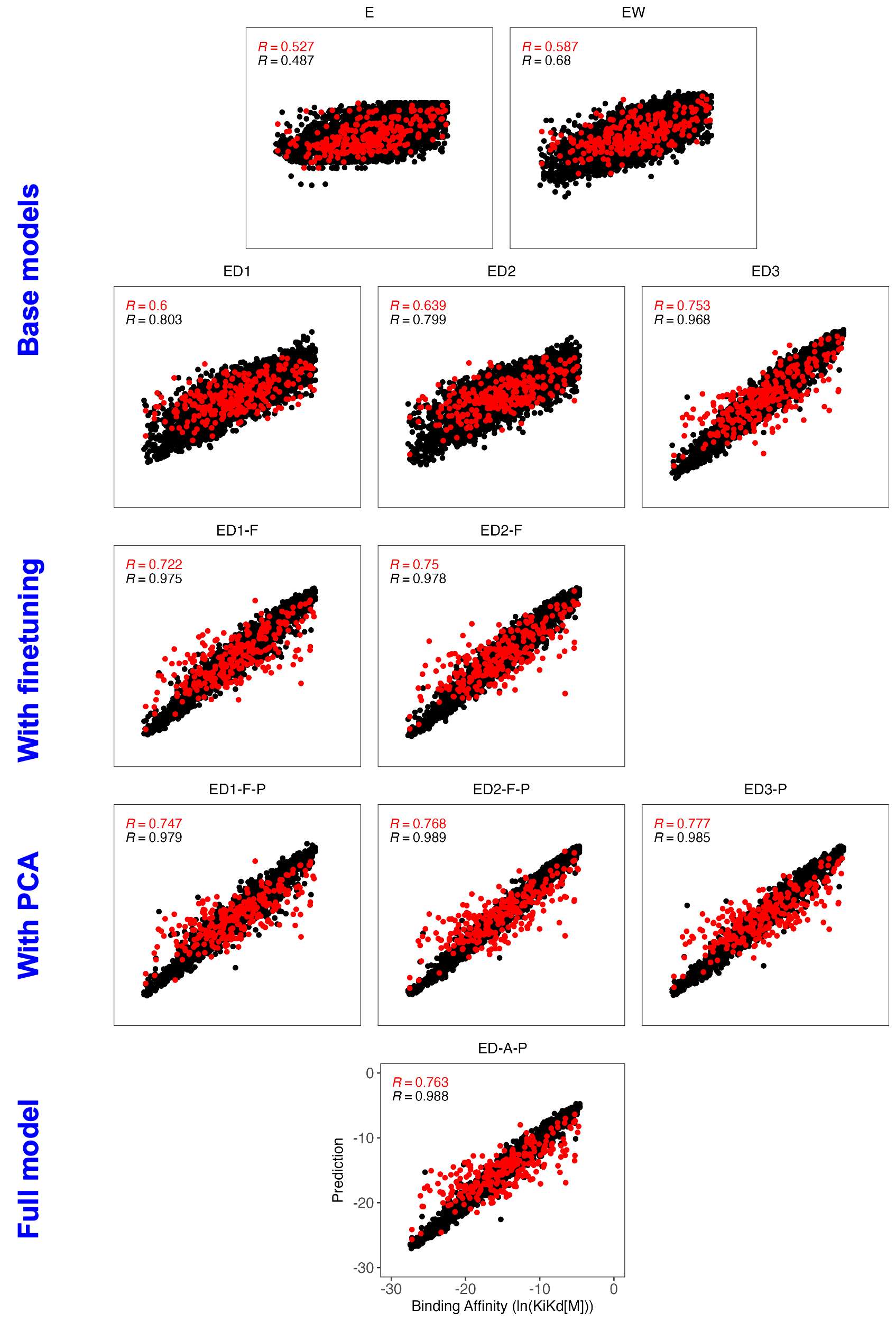}}
\par\bigskip
\raggedright\textbf{Figure 5. Performances of meta-models.} Predictions by XGBoost for the unfiltered training set are shown. The
first row represents the meta-models constructed entirely based on the docking methods, SMINA and Vinardo. The first
panel shows results for the two scores alone (E), while the second panel shows the results also including the molecular
weight parameter (EW). The remaining rows show results for meta-models that include DL scores. The second, third and
fourth rows represent successive processing of the deep learning model scores, with the second row showing the average
scores (ED1, ED2, ED3), the third row (if a panel is present) showing the average after fine-tuning (F) using the
RefinedSet{\textbackslash}CoreSet (ED1-F, ED2-F), and the fourth row showing the scores using the optimal number of PCs
from a PCA (P) on the previous level (either fine-tuned or not) (ED1-F-P, ED2-F-P, ED3-P). The columns of the second,
third and fourth rows show the three different base training sets used for the DL scores, where the leftmost column is
based on the pre-trained models on BindingDB (ED1 series), the central column is based on the de novo-trained models on
BindingDB (ED2 series) and the rightmost column is based on the de novo-trained models on PDBbind (ED3 series). The
single panel in the last fifth row shows the results for the meta-model ED-A-P, which is based on a PCA on all the
training scores from ED1-F, ED2-F, and ED3.

\clearpage\centering
\lfbox[margin-bottom=0.0028in,margin-top=0mm,margin-right=0mm,margin-left=0mm,border-style=none,padding=0mm]{\includegraphics[width=6in,height=5.9563in]{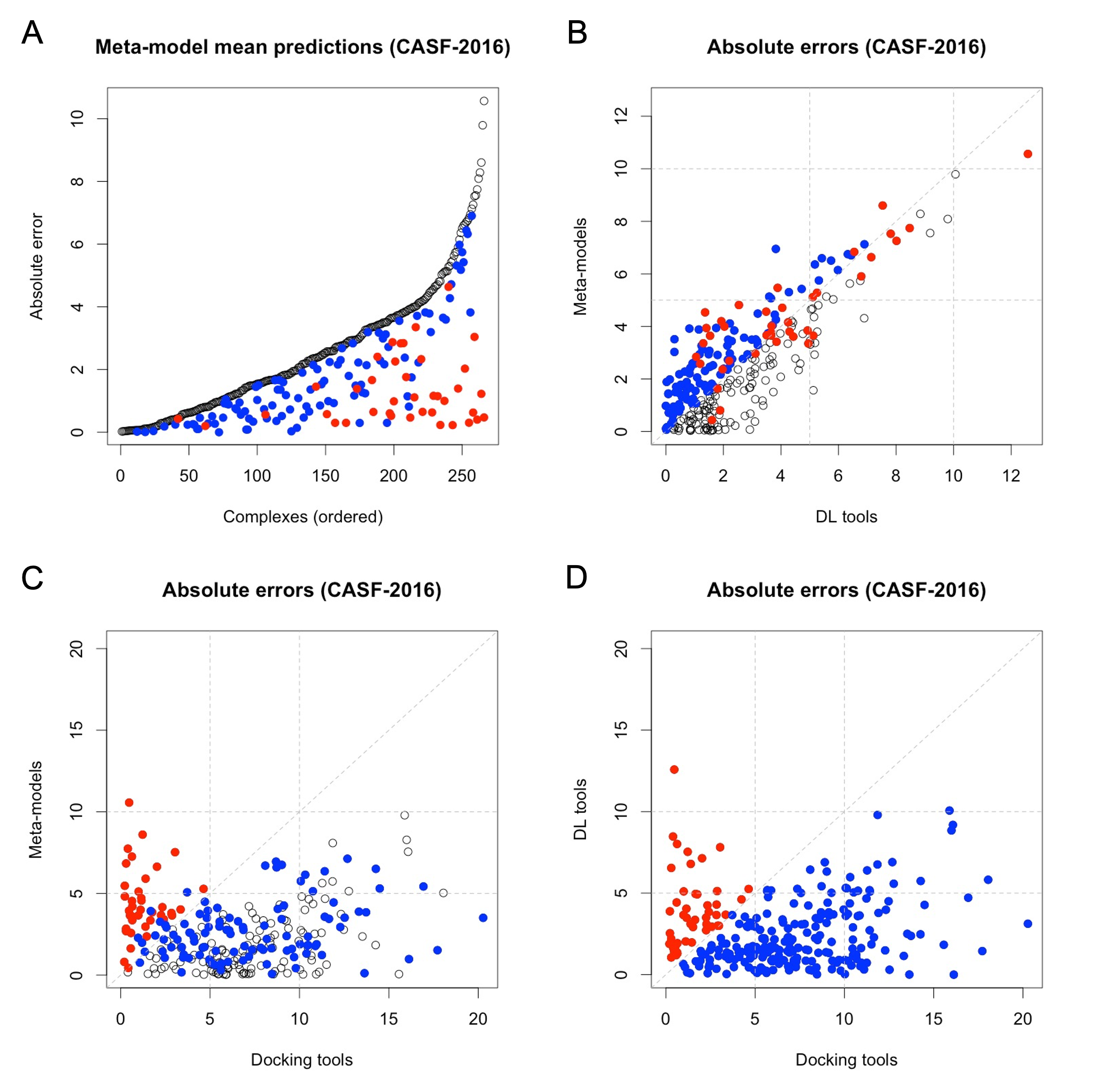}}
\par\bigskip
\raggedright\textbf{Figure 6. Prediction synergy and complementarity of docking and DL tools by meta-modeling.} (A) Absolute errors (AEs) of the mean predictions for the CASF-2016 benchmark set (266 complexes) by the 4 meta-models with ED-A-P (open
circles; ordered by AEs). The blue circles correspond to 104 complexes (39.1\%) whose predictions by the DL tools
(DAP{\textbar}PC1 in Fig. 4D) are better than the mean predictions by the 4 meta-models or the 4 docking tools (i.e.,
lower AEs). The red circles correspond to 39 complexes (14.7\%) whose mean predictions by the 4 docking tools are
better than those by the meta-models or the DAP{\textbar}PC1. (B-D) Scatter plots of AEs of the (mean) predictions for
the CASF-2016 benchmark set by the DAP{\textbar}PC1 vs. the 4 meta-models (B), by the 4 docking tools vs. the 4
meta-models (C), and by the 4 docking tools vs. the DAP{\textbar}PC1 (D). The blue and red dots in (B) and (C)
correspond to those in (A). In (D), the 48 complexes (18.0\%) in red indicate better predictions (i.e., lower AEs) by
the docking tools on average, whereas the remaining 218 complexes (82.0\%) in blue indicate better predictions by the
DAP{\textbar}PC1.


\newpage

\clearpage\section[Supporting Information]{\textbf{Supporting Information}}
\section[Improved Prediction Of Ligand{}-Protein Binding Affinities By Meta{}-Modeling]{Improved Prediction Of
Ligand-Protein Binding Affinities By Meta-Modeling}
Ho-Joon Lee\textsuperscript{1,2,*\#}, Prashant S. Emani\textsuperscript{3,*\#} and Mark B.
Gerstein\textsuperscript{3,4,5,6,7\#}

\textsuperscript{1}Dept. of Genetics and \textsuperscript{2}Yale Center for Genome Analysis, Yale University, New Haven,
CT 06510, USA\par
\textsuperscript{3}Dept. of Molecular Biophysics \& Biochemistry, Yale University, New Haven, CT 06520, USA\par
\textsuperscript{4}Program in Computational Biology \& Bioinformatics, \textsuperscript{5}Dept. of Computer Science,
\textsuperscript{6}Dept. of Statistics \& Data Science, and \textsuperscript{7}Dept. of Biomedical Informatics \& Data
Science, Yale University, New Haven, CT 06520, USA
\bigskip

\raggedright
* Equal contributions\par
\# Correspondence: HL, \href{mailto:ho-joon.lee@yale.edu}{\textcolor{blue}{ho-joon.lee@yale.edu}}; PSE,
\href{mailto:prashant.emani@yale.edu}{\textcolor{blue}{prashant.emani@yale.edu}}; MBG,
\href{mailto:pi@gersteinlab.org}{\textcolor[HTML]{1155CC}{pi@gersteinlab.org}}

\clearpage
\subsection{Supplementary Methods}
\subsubsection{Docking Methods}
We have constructed custom Python wrapper scripts to implement the entire pipeline, with slight variations depending on
the dataset considered. We describe the essential aspects of running a docking simulation in the following.

\begin{enumerate}[series=listWWNumxxx,label=\textstyleListLabelccxxxi{\arabic*.},ref=\arabic*]
\item \textit{Selection of the receptor structure file:} All data files for each complex in the PDBbind dataset
\textsuperscript{1} are contained in separate subfolders. We utilized the protein PDB file for the remaining processing
steps.

\item \textit{Preparation of the receptor structure:} The receptor structure is read in as a PDB file and preprocessed
in PyMol (\url{https://pymol.org}). We used the \textit{pymol} class and associated functions within our pipeline
script. The preprocessing consists of the removal of \textit{all }solvent and heteroatoms in the molecules. The
resulting PDB file is then converted into the requisite PDBQT format using OpenBabel (\url{http://openbabel.org}). We
used the \textit{obabel} flags $\text{\textgreek{‘}}$-e = continue after errors$\text{\textgreek{’}}$,
$\text{\textgreek{‘}}$-p = add hydrogens appropriate for the pH$\text{\textgreek{’}}$, and
$\text{\textgreek{‘}}$-{}-partialcharge gasteiger$\text{\textgreek{’}}$ to calculate Gasteiger partial charges. The pH
was set at 7.0. We included the PDBQT-specific flags $\text{\textgreek{‘}}$r = output as a rigid
molecular$\text{\textgreek{’}}$ and $\text{\textgreek{‘}}$p = Preserve atom indices from input
file$\text{\textgreek{’}}$. The implementation of this is written as

\hspace*{0.5cm} obabel -ipdb {\textless}Input PDB file{\textgreater} -opdbqt -O {\textless}Output PDBQT file{\textgreater} -p 7.0 -e
-xrp -{}-partialcharge gasteiger

(Note the $\text{\textgreek{‘}}$-x$\text{\textgreek{’}}$ flag is set to enable the output-specific flags, which are
PDBQT-specific in our case.)

\item \textit{Ligand files:} Given the use of 3D SDF files for all analyses, the ligand files did not need further
processing. The appropriate ligand SDF files were invoked in the docking command (see below), including the
complex-specific ligand files in the PDBbind dataset (the ligand SDF files are included in the subfolders for each
complex).

\item \textit{Creation of the configuration file:} The configuration file serves to provide a set of commands to the
\textit{smina} package \textsuperscript{2} (as well as the original AutoDock Vina tool \textsuperscript{3}). The file
contains the receptor PDBQT file, the ligand SDF file, the center coordinates of the search box in the receptor
structure file coordinate system, and the dimensions of the search box. The calculation of the center coordinates
varies depending on the datasets under consideration. For the PDBbind dataset, we used the center of mass of each
ligand SDF file as the center of the search box, employing PyMol$\text{\textgreek{’}}$s \textit{centerofmass} function
to do so. The search box was set to have dimensions of 40 Angstroms in the x, y and z directions.

\item \textit{Running the docking software:} Using the configuration file and the ligand file name, we ran the program
\textit{smina} using both the default (hereafter termed “SMINA”) and “Vinardo” scoring functions. We used an
\textit{exhaustiveness} parameter of 32 and used the default output of the top 9 poses (ranked from 0 to 8 from lowest
to highest binding energy). The program call was: 

\hspace*{0.5cm} smina -{}-config {\textless}config\_file{\textgreater} -{}-out {\textless}Output Complex PDBQT file{\textgreater}
-{}-log {\textless}Log file containing binding affinities{\textgreater} -{}-exhaustiveness 32
\end{enumerate}

\subsubsection{Implementation of RMSD filters}
For the RMSD filters, we calculate  $\mathit{RMSD}_{\mathit{ab}}$ \ in two ways, depending on the choice of docking
poses:

\begin{enumerate}[series=listWWNumxxix,label=\textstyleListLabelccxli{\alph*.},ref=\alph*]
\item \textbf{Experimental pose-RMSD filtering:} The deviation is calculated for a ligand pose predicted by a docking
method relative to the experimental structure of the ligand. Ligands were assigned an RMSD of 100 Angstroms if either
of the docking tools failed to generate a docked structure, or if the bond reordering process (see
\textbf{Supplementary Methods}) failed. In those cases, we automatically selected the lowest energy pose out of the 9
outputs by the docking tools. After reordering, the symmetric RMSD was calculated, and we filtered the poses for each
complex by an RMSD cutoff of 3 Angstroms. Among those poses that satisfied the cutoff we chose the lowest energy
structure. If no poses met the cutoff, then we selected the lowest energy pose with the RMSD set to 100 Angstroms. In
this way, we had two tiers of poses from all the complexes: a set of docked poses with RMSD relative to the
experimental poses {\textless} 3 Angstroms (and which may not have been the lowest energy pose among the 9 docking
poses); and another set where the lowest energy pose was selected with an arbitrarily set RMSD of 100 Angstroms (for
downstream filtering purposes).
\item \textbf{Consensus pose-RMSD filtering:} This is where the deviation is calculated between the docking poses for
the same ligand-target pair as predicted by two different docking methods. In our analyses, these correspond to the
poses from the SMINA and Vinardo scoring functions. For the Vinardo and SMINA pose comparisons, we had 9  $\times $ \ 9
= 81 pairs of docked poses to filter through for the final assignment of RMSD. In the case of docked pose comparisons,
there was no need for bond reordering, as \textit{smina} generated poses with the same ordering convention using both
scoring functions. The RMSD was calculated for all pairs of docking poses for each complex and the [SMINA, Vinardo]
pair with an RMSD less than 3 Angstroms and with the lowest rank sum (SMINA rank + Vinardo rank) was chosen in the
final scoring. If multiple pairs had the same rank sum, the pair with the lowest RMSD was selected. If a complex had no
pairs with an RMSD less than 3 Angstroms, then the lowest energy poses from both methods were selected and assigned a
pairwise RMSD of 100 Angstroms for downstream filtering. As a result of this assignment process, all structures with an
RMSD {\textless} 100 Angstroms will necessarily have an RMSD {\textless} 3 Angstroms. We thus only need to consider one
of the two cases in the results.
\end{enumerate}

Note on the coding of \textit{Experimental pose-RMSD filtering}: The experimental structures
are available, in our analyses, in the PDBbind dataset in SDF and MOL2 formats. The calculation of this RMSD in
Pymol$\text{\textgreek{’}}$s Python API (\url{http://www.pymol.org/pymol}) involved first reordering the bonds in the
docking pose using either the experimental structure or the SMILES string as a “template”, as the atoms are not
consistently designated between the \textit{smina} predicted poses and the experimental structures. This was followed
by the calculation of the RMSD between the experimental and reordered predicted pose. The bond reordering process was
carried out using the experimental mol2 file (for the PDBbind structures), using functions in the Python package RDKit
(\href{http://www.rdkit.org/}{\textcolor[HTML]{1155CC}{www.rdkit.org/}}).

\subsubsection[Meta{}-models]{Meta-models}
We used linear regression, ElasticNet, and LASSO algorithms for linear meta-models and the XGBoost algorithm for a
non-linear meta-model implemented in \textit{scikit-learn} in Python \textsuperscript{4}. The steps in the
preprocessing of the PDBbind dataset were as follows:

\begin{enumerate}[series=listWWNumxxiv,label=\textstyleListLabelcxcv{\arabic*.},ref=\arabic*]
\item The predicted binding affinity matrices from the docking and deep learning tools were merged based on the PDB IDs.
\item We separately incorporated the RMSD values for the complexes (either based on
$\text{\textgreek{‘}}$Experimental$\text{\textgreek{’}}$ or $\text{\textgreek{‘}}$Consensus$\text{\textgreek{’}}$
quantifications) into the merged binding affinity matrix.
\item Those structures in the non-core-set not satisfying the RMSD cutoffs were removed from further analysis. We did
not apply the RMSD filters to the core set. 
\item We also calculated the molecular weight of the ligands and allowed for the possibility of incorporating these
values into the binding affinity matrix (this is enabled by a separate flag in the code, which either includes or
excludes the molecular weight).
\item All structures in the core set of the PDBbind cohort are set aside for the test scoring. The remainder of the
binding affinity matrix is carried forward for the analysis.
\item The binding affinity matrix for the training set (the non-core-set portion of the PDBbind refined set) was
separated into feature (X) and target value (y) numpy arrays.
\end{enumerate}
\bigskip

These trainsets were input to the meta-models. The parameters used in each of the models are:

\begin{enumerate}[series=listWWNumxxv,label=\textstyleListLabelcciv{\Alph*.},ref=\Alph*]
\item \textbf{LASSO model:} We iterated \textit{scikit-learn}$\text{\textgreek{’}}$s LassoCV approach 100 times with
shuffled data, using 5-fold cross-validation each time. The best model chosen was the one with the minimum
$\text{\textgreek{‘}}$dual\_gap\_$\text{\textgreek{’}}$ score.
\item \textbf{ElasticNet model:} We iterated \textit{scikit-learn}$\text{\textgreek{’}}$s ElasticNetCV approach 10 times
for each L1\_ratio in the list [.1, .5, .7, .9, .95, .99, 1] (the L1\_ratio parameter is the balance between the L1 and
L2 penalties in the ElasticNet method). Each iteration used shuffled data and 5-fold cross-validation. The best model
chosen overall was the one with the minimum $\text{\textgreek{‘}}$dual\_gap\_$\text{\textgreek{’}}$ score.
\item \textbf{Linear Regression model:} We ran linear regression with no additional parameters or regularization
penalties.
\item \textbf{XGBoost model:} We used the \textit{xgboost.XGBRegressor} model class with a “squared-error” objective
(\url{https://xgboost.readthedocs.io/en/release_0.72/python}). This model was then used in
\textit{scikit-learn}$\text{\textgreek{’}}$s RandomizedSearchCV method to run 100 iterations with 5-fold
cross-validation, comprehensively exploring several parameters within reasonable ranges:
{\textquotedbl}colsample\_bytree{\textquotedbl} uniformly in the range [0.8, 0.2]; {\textquotedbl}gamma{\textquotedbl}
uniformly in the range [0, 0.5]; {\textquotedbl}learning\_rate{\textquotedbl} uniformly in the range [0.02, 0.3];
{\textquotedbl}max\_depth{\textquotedbl} randomly from the set of integers in the range [2, 6];
{\textquotedbl}n\_estimators{\textquotedbl} randomly from the set of integers in the range [100, 150];
{\textquotedbl}subsample{\textquotedbl} uniformly in the range [0.7, 0.3]. The maximum
$\text{\textgreek{‘}}$mean\_test\_score$\text{\textgreek{’}}$ was used to select the best model.
\end{enumerate}

\subsubsection{SwissADME Analysis}
\textcolor{black}{To understand the contributions of ligand properties to performance of all the models, we used the web
tool }\textit{\textcolor{black}{SwissADME}}\textcolor{black}{ }\textcolor{black}{\textsuperscript{5}}\textcolor{black}{
to extract the properties of the PDBbind CoreSet ligands from their SMILES strings. We first converted the ligand
SMILES strings to canonical SMILES format using OpenBabel (}\url{http://openbabel.org}\textcolor{black}{). We then
submitted the canonical SMILEs strings to SwissADME$\text{\textgreek{’}}$s web interface. This yielded a set of 46
properties for 263 ligands in the CoreSet. We subsequently sought to measure the correlation between the deviation of
the predicted scores from the experimental values with the numerical features, to identify those ligand features that
might contribute to reduced performance of the model.}
\bigskip

We also investigated the SwissADME features that might be associated with better or worse performance for subsets of
complexes by the different groups of tools (docking, DL, and meta-models). We checked for duplicate SMILES strings for
the ligands and found none, thus requiring no further filtration. Using two-sided Wilcoxon and t-tests, we compared the
distributions for all the numeric SwissADME features.

\subsubsection{UniProt Analysis}
To identify protein features that might be associated with better or worse performance of specific tools, we extracted
the relevant subsets of PDBbind CoreSet PDB IDs from our analyses and fed them into the “Retrieve/ID mapping” tool of
the UniProt database (\href{http://www.uniprot.org/id-mapping}{\textcolor[HTML]{1155CC}{www.uniprot.org/id-mapping}})
\textsuperscript{6}. The corresponding UniProt IDs and the following features were downloaded in a .tsv file: 'From',
'Entry', 'Reviewed', 'Entry Name', 'Protein names', 'Gene Names', 'Organism', 'Length', 'Active site', 'Binding site',
'Function [CC]', 'Catalytic activity', 'Annotation', 'Gene Ontology (biological process)', 'Helix', 'Turn', 'Beta
strand', 'Coiled coil', 'Compositional bias', 'Domain [CC]', 'Domain [FT]', 'Motif', 'Protein families', 'Region',
'Repeat', 'Zinc finger', 'Pathway', 'Gene Ontology (molecular function)', 'Post-translational modification'. We first
filtered out all UniProt entries that were not labeled “reviewed”. Next, we manually checked repeats, where the same
PDB ID mapped to multiple UniProt IDs, by looking up the PDB IDs on the RCSB PDB website
(\href{http://www.rcsb.org/}{\textcolor[HTML]{1155CC}{www.rcsb.org/}}). We checked both the “Gene Names” and “Species”
and removed all UniProt entries that did not match the PDB entry. In all cases, this left only a single PDB ID-UniProt
ID mapping for each PDB ID. Additionally, to run the tests of over- or under-representation, we also deduplicated the
UniProt IDs. That is, we removed all cases of proteins that were represented more than once, as the CoreSet has a large
number of proteins that recur in the dataset. This led to a significant reduction in the number of proteins tested for
UniProt feature prioritization from 266 to 60. Fisher$\text{\textgreek{’}}$s exact test was run when comparing the
number of occurrences of a feature in the subset versus the whole. The features tested are: the number of times a given
UniProt term was annotated at all; the occurrence of Gene Ontology \textsuperscript{7} terms, both the “Biological
Process” and “Molecular Function” categories; and protein Domain and Pathway annotations.

\subsubsection[LIT{}-PCBA Virtual Screening]{LIT-PCBA Virtual Screening}
For the LIT-PCBA virtual screening exercise, we utilized sets of active and inactive ligands in the ratio of 1:2
randomly drawn from the combined training and validation cohorts of the LIT\_PCBA database\textsuperscript{8} (see
\textbf{Methods}). To run the docking tools on the datasets, we generated three-dimensional SDF structures for each of
the ligands from their SMILES strings using the following command:

\hspace*{.5cm} obabel -:'{\textless}SMILES string{\textgreater}' -{}-gen2d -omol {\textbar} obabel -imol -{}-minimize -{}-noh -omol
{\textbar} obabel -imol -h -e -{}-gen3d -osdf -O '{\textless}output SDF file{\textgreater}'
\bigskip

We also utilized the LIT-PCBA-provided PDB structures for the protein-ligand complex to configure the docking, by
removing solvent molecules and other heteroatoms from the mol2 structure of the protein and then using the
center-of-mass coordinates of the corresponding docked ligand in the PDB complex to define the docking search box. This
facilitates a search for poses within the same binding pocket as the template complex provided by LIT-PCBA.

\clearpage
\subsection{Supplementary References}\par

1.\ \ Liu, Z.; \ Su, M.; \ Han, L.; \ Liu, J.; \ Yang, Q.; \ Li, Y.; Wang, R., Forging the Basis for Developing
Protein–Ligand Interaction Scoring Functions. \textit{Accounts of Chemical Research }\textbf{2017,} \textit{50} (2),
302-309.
\vskip 0.1in
2.\ \ Koes, D. R.; \ Baumgartner, M. P.; Camacho, C. J., Lessons learned in empirical scoring with smina from the CSAR
2011 benchmarking exercise. \textit{J Chem Inf Model }\textbf{2013,} \textit{53} (8), 1893-904.
\vskip 0.1in
3.\ \ Trott, O.; Olson, A. J., AutoDock Vina: improving the speed and accuracy of docking with a new scoring function,
efficient optimization, and multithreading. \textit{J Comput Chem }\textbf{2010,} \textit{31} (2), 455-61.
\vskip 0.1in
4.\ \ Pedregosa, F.; \ Varoquaux, G.; \ Gramfort, A.; \ Michel, V.; \ Thirion, B.; \ Grisel, O.; \ Blondel, M.;
\ Prettenhofer, P.; \ Weiss, R.; \ Dubourg, V.; \ Vanderplas, J.; \ Passos, A.; \ Cournapeau, D.; \ Brucher, M.;
\ Perrot, M.; Duchesnay, É., Scikit-learn: Machine Learning in Python. \textit{J. Mach. Learn. Res. }\textbf{2011,}
\textit{12} (null), 2825–2830.
\vskip 0.1in
5.\ \ Daina, A.; \ Michielin, O.; Zoete, V., SwissADME: a free web tool to evaluate pharmacokinetics, drug-likeness and
medicinal chemistry friendliness of small molecules. \textit{Scientific reports }\textbf{2017,} \textit{7} (1), 42717.
\vskip 0.1in
6.\ \ The UniProt Consortium, UniProt: the Universal Protein Knowledgebase in 2023. \textit{Nucleic acids research
}\textbf{2023,} \textit{51} (D1), D523-d531.
\vskip 0.1in
7.\ \ Aleksander, S. A.; \ Balhoff, J.; \ Carbon, S.; \ Cherry, J. M.; \ Drabkin, H. J.; \ Ebert, D.; \ Feuermann, M.;
\ Gaudet, P.; \ Harris, N. L.; \ Hill, D. P.; \ Lee, R.; \ Mi, H.; \ Moxon, S.; \ Mungall, C. J.; \ Muruganugan, A.;
\ Mushayahama, T.; \ Sternberg, P. W.; \ Thomas, P. D.; \ Van Auken, K.; \ Ramsey, J.; \ Siegele, D. A.; \ Chisholm, R.
L.; \ Fey, P.; \ Aspromonte, M. C.; \ Nugnes, M. V.; \ Quaglia, F.; \ Tosatto, S.; \ Giglio, M.; \ Nadendla, S.;
\ Antonazzo, G.; \ Attrill, H.; \ Dos Santos, G.; \ Marygold, S.; \ Strelets, V.; \ Tabone, C. J.; \ Thurmond, J.;
\ Zhou, P.; \ Ahmed, S. H.; \ Asanitthong, P.; \ Luna Buitrago, D.; \ Erdol, M. N.; \ Gage, M. C.; \ Ali Kadhum, M.;
\ Li, K. Y. C.; \ Long, M.; \ Michalak, A.; \ Pesala, A.; \ Pritazahra, A.; \ Saverimuttu, S. C. C.; \ Su, R.;
\ Thurlow, K. E.; \ Lovering, R. C.; \ Logie, C.; \ Oliferenko, S.; \ Blake, J.; \ Christie, K.; \ Corbani, L.;
\ Dolan, M. E.; \ Drabkin, H. J.; \ Hill, D. P.; \ Ni, L.; \ Sitnikov, D.; \ Smith, C.; \ Cuzick, A.; \ Seager, J.;
\ Cooper, L.; \ Elser, J.; \ Jaiswal, P.; \ Gupta, P.; \ Jaiswal, P.; \ Naithani, S.; \ Lera-Ramirez, M.; \ Rutherford,
K.; \ Wood, V.; \ De Pons, J. L.; \ Dwinell, M. R.; \ Hayman, G. T.; \ Kaldunski, M. L.; \ Kwitek, A. E.;
\ Laulederkind, S. J. F.; \ Tutaj, M. A.; \ Vedi, M.; \ Wang, S. J.; \ D'Eustachio, P.; \ Aimo, L.; \ Axelsen, K.;
\ Bridge, A.; \ Hyka-Nouspikel, N.; \ Morgat, A.; \ Aleksander, S. A.; \ Cherry, J. M.; \ Engel, S. R.; \ Karra, K.;
\ Miyasato, S. R.; \ Nash, R. S.; \ Skrzypek, M. S.; \ Weng, S.; \ Wong, E. D.; \ Bakker, E.; \ Berardini, T. Z.;
\ Reiser, L.; \ Auchincloss, A.; \ Axelsen, K.; \ Argoud-Puy, G.; \ Blatter, M. C.; \ Boutet, E.; \ Breuza, L.;
\ Bridge, A.; \ Casals-Casas, C.; \ Coudert, E.; \ Estreicher, A.; \ Livia Famiglietti, M.; \ Feuermann, M.; \ Gos, A.;
\ Gruaz-Gumowski, N.; \ Hulo, C.; \ Hyka-Nouspikel, N.; \ Jungo, F.; \ Le Mercier, P.; \ Lieberherr, D.; \ Masson, P.;
\ Morgat, A.; \ Pedruzzi, I.; \ Pourcel, L.; \ Poux, S.; \ Rivoire, C.; \ Sundaram, S.; \ Bateman, A.;
\ Bowler-Barnett, E.; \ Bye, A. J. H.; \ Denny, P.; \ Ignatchenko, A.; \ Ishtiaq, R.; \ Lock, A.; \ Lussi, Y.;
\ Magrane, M.; \ Martin, M. J.; \ Orchard, S.; \ Raposo, P.; \ Speretta, E.; \ Tyagi, N.; \ Warner, K.; \ Zaru, R.;
\ Diehl, A. D.; \ Lee, R.; \ Chan, J.; \ Diamantakis, S.; \ Raciti, D.; \ Zarowiecki, M.; \ Fisher, M.; \ James-Zorn,
C.; \ Ponferrada, V.; \ Zorn, A.; \ Ramachandran, S.; \ Ruzicka, L.; Westerfield, M., The Gene Ontology knowledgebase
in 2023. \textit{Genetics }\textbf{2023,} \textit{224} (1).
\vskip 0.1in
8.\ \ Tran-Nguyen, V.-K.; \ Jacquemard, C.; Rognan, D., LIT-PCBA: An Unbiased Data Set for Machine Learning and Virtual
Screening. \textit{Journal of Chemical Information and Modeling }\textbf{2020,} \textit{60} (9), 4263-4273.


\newpage

\section{Supplementary Figures}

\raggedright\textbf{Figure S1. Overview of our workflow with an example meta-model of ED1-F-P.}\par
\centering\includegraphics[width=6.5in,height=6.7244in]{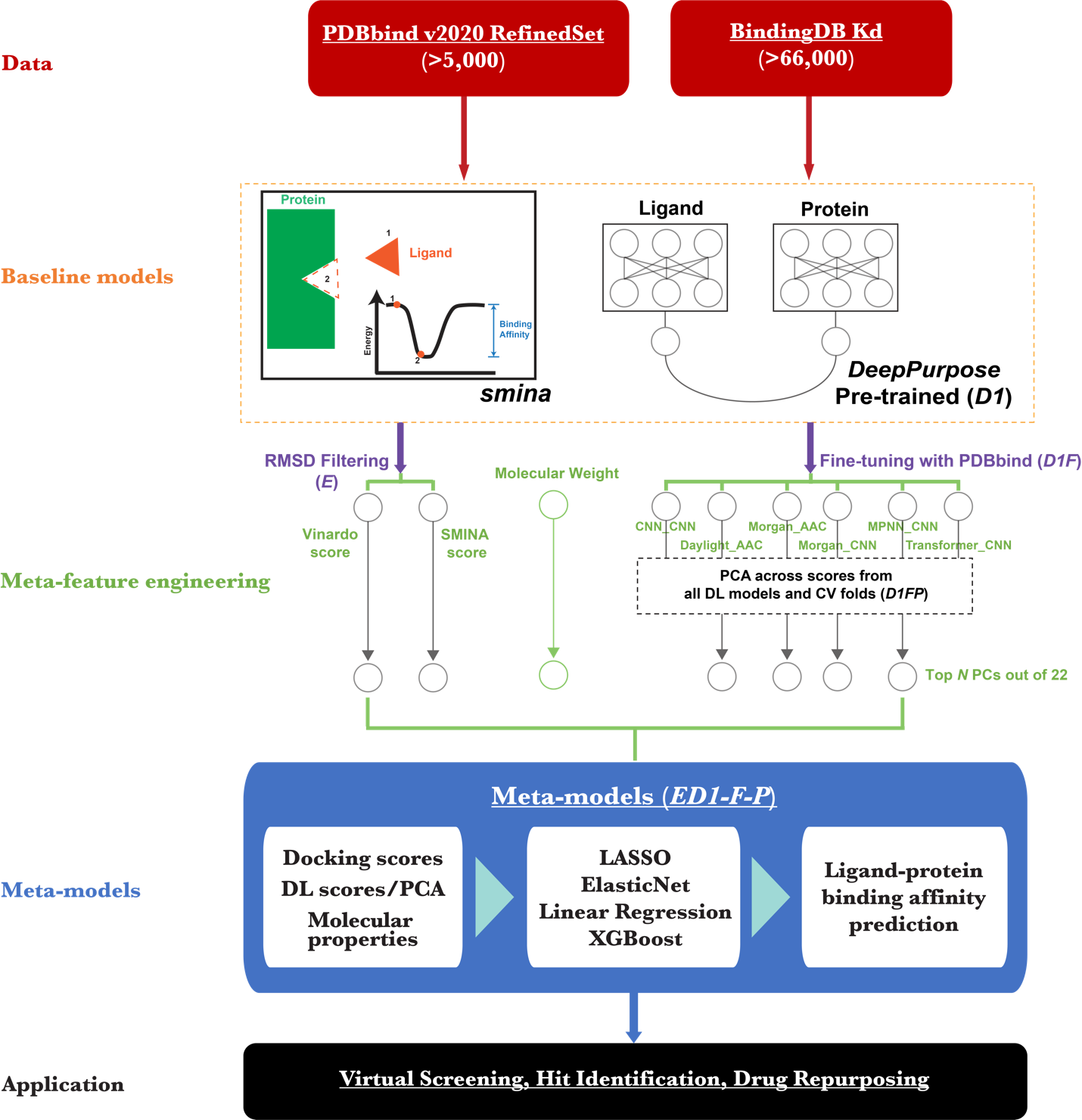}

\clearpage
\raggedright\textbf{Figure S2. Ligand pose-based filtering for the docking tools. }(A) RMSD distributions for the experimental
pose-based and consensus pose-based filtering strategies (266 CASF-2016 complexes in red) (B) Monte Carlo simulations
of prediction correlations for the CoreSet 3A filtering.\par
\centering\includegraphics[width=6.5in,height=6.4339in]{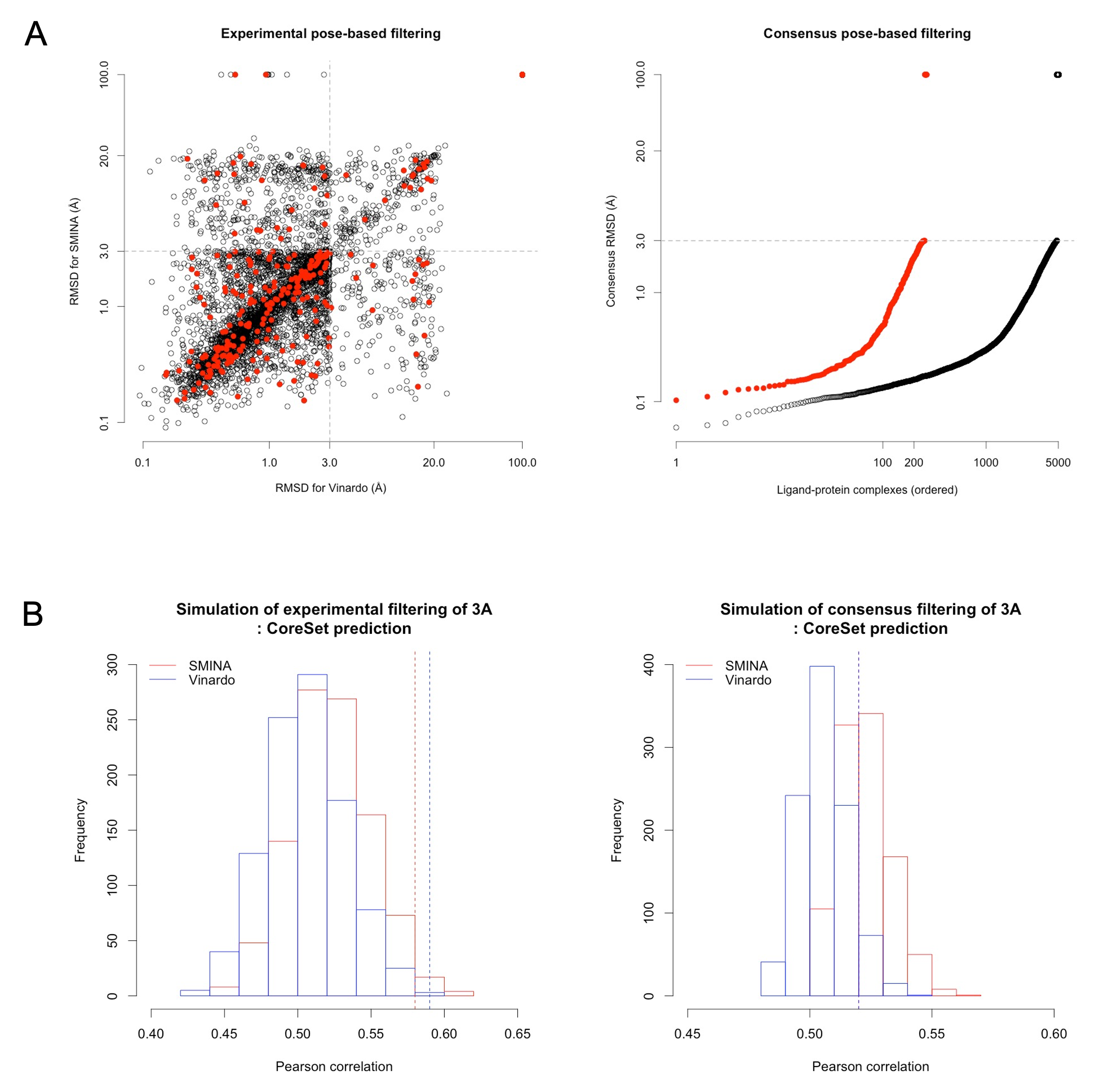}

\clearpage
\raggedright\textbf{Figure S3. Training and validation of 12 initial BindingDB-trained deep learning models.}\par
\centering\includegraphics[width=6.5138in,height=4.3772in]{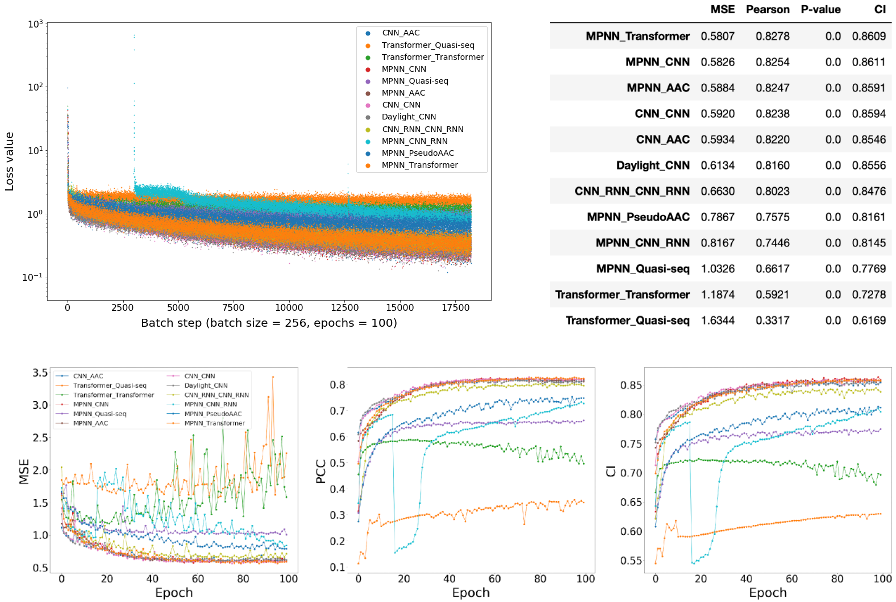}

\clearpage
\raggedright\textbf{Figure S4.} \textbf{Regression results of the BindingDB Kd data for 66,444 ligand-protein complexes by the
BindingDB-trained, PDBbind-trained, and PDBbind-finetuned models.}\par
\bigskip
\centering\includegraphics[width=6.5in,height=3.7835in]{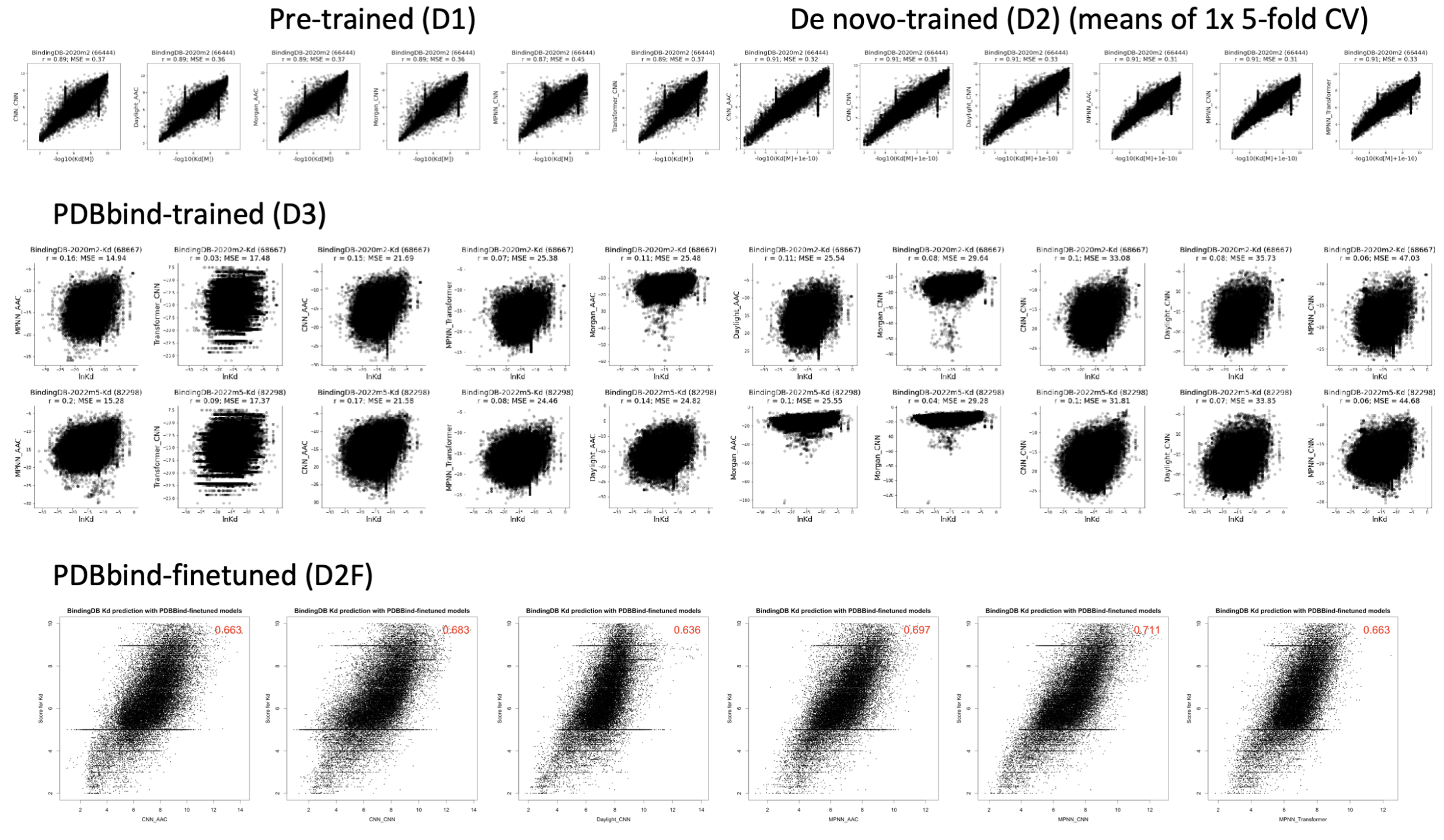} 

\clearpage
\raggedright\textbf{Figure S5.} \textbf{Regression results of the PDBBind-v2020-RefinedSet by the BindingDB pre-trained, BindingDB}\textbf{\textit{de novo}}\textbf{{}-trained, PDBbind }\textbf{\textit{de novo}}\textbf{{}-trained, and PDBbind-finetuned models.}\par
\bigskip
\centering\includegraphics[width=6.5in,height=4.3654in]{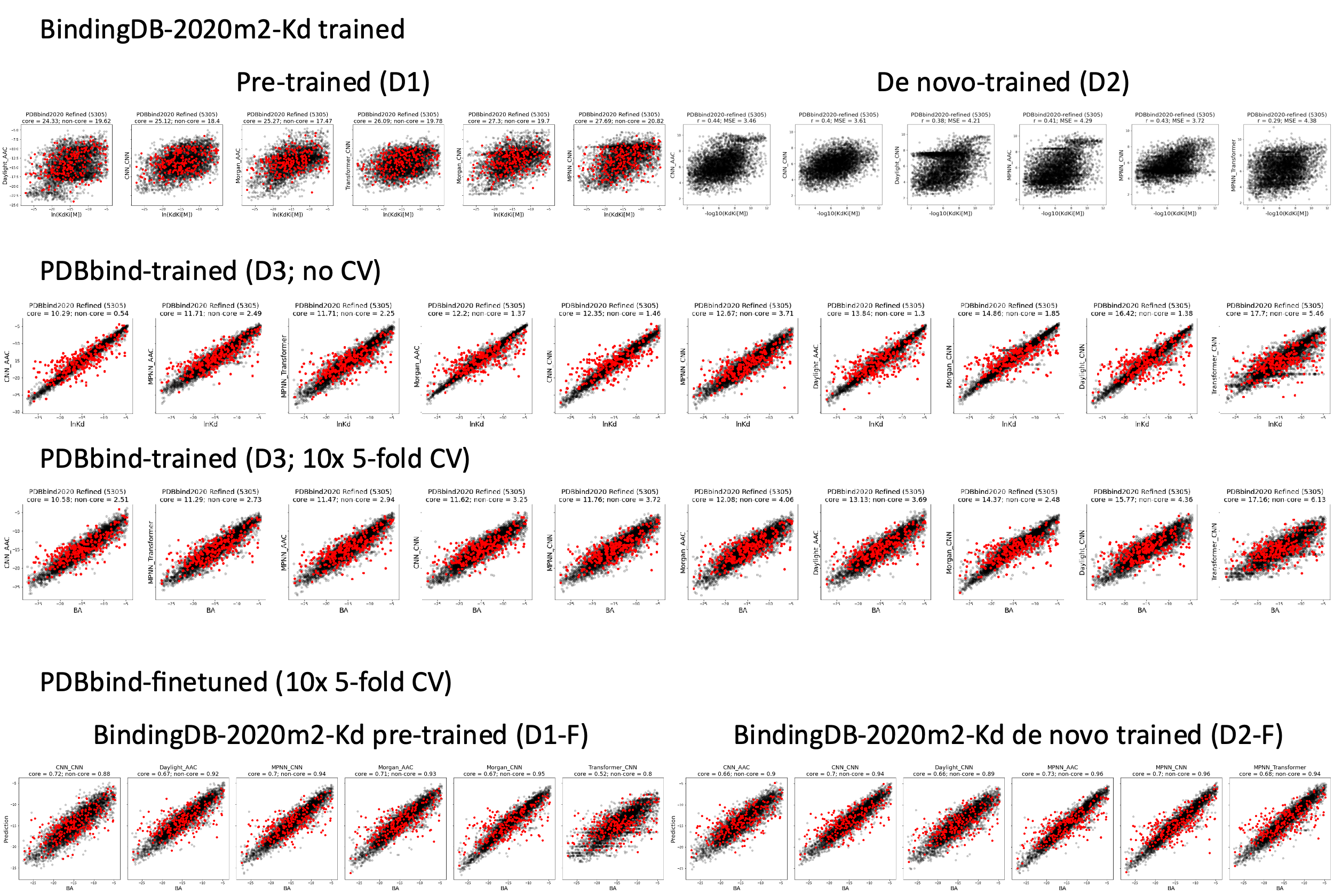} 

\clearpage
\raggedright\textbf{Figure S6.} \textbf{Model comparison of non-core (training set) predictions.}\par
\lfbox[margin-bottom=0.0063in,margin-top=0mm,margin-right=0mm,margin-left=0mm,border-style=none,padding=0mm]\centering\includegraphics[width=6.5in,height=6.7992in]{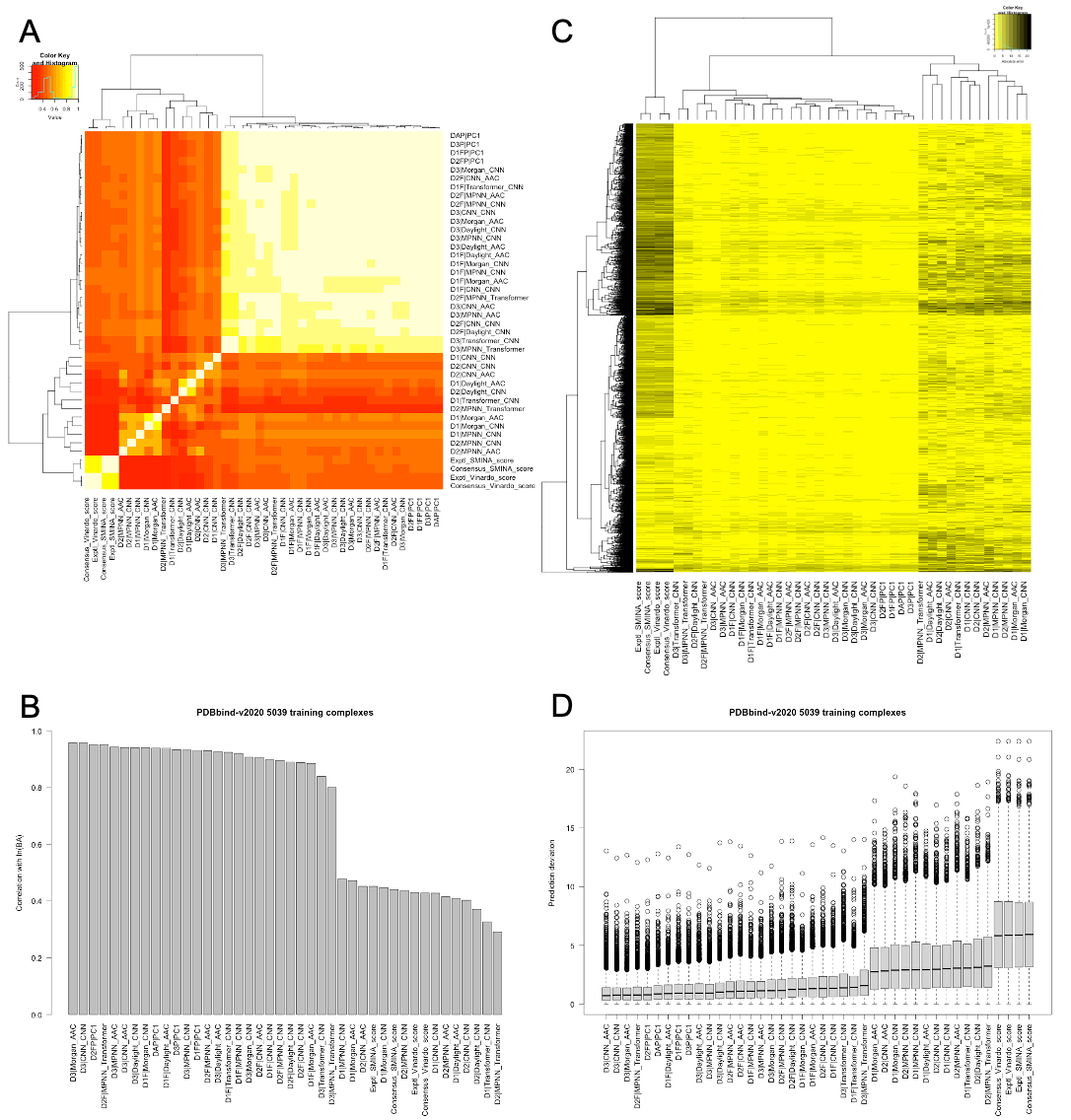}

\clearpage
\raggedright\textbf{Figure S7. Feature importance of meta-models.}\par
\centering \includegraphics[width=6.5in,height=7.1528in]{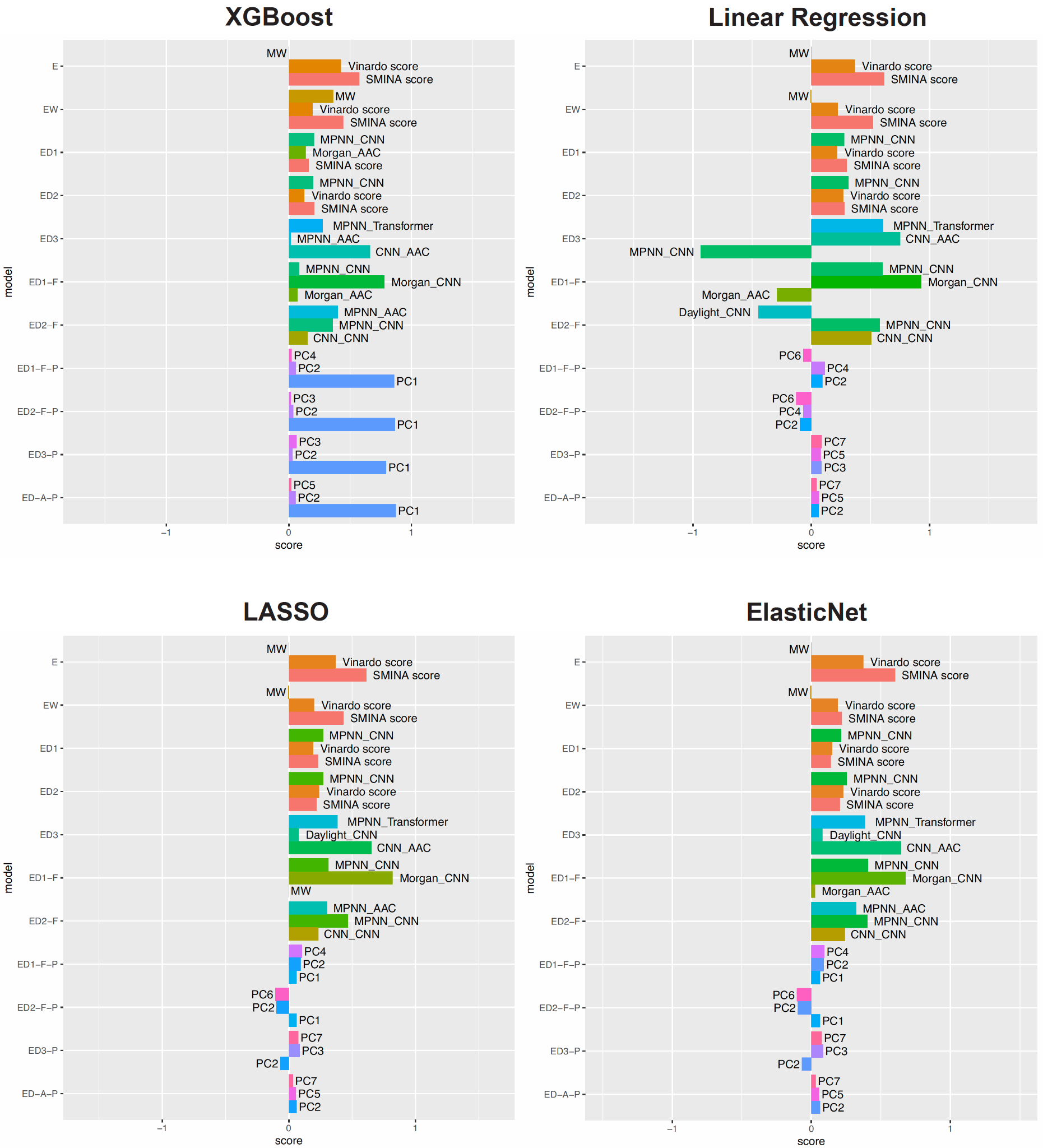}

\clearpage
\raggedright\textbf{Figure S8. Performance of meta-models for PBDbind2020 GeneralSet as a secondary benchmark}. Predictions by XGBoost with the Consensus filtering and no RMSD filter are shown.\par
\bigskip
\centering\includegraphics[width=5.0in,height=7.5in]{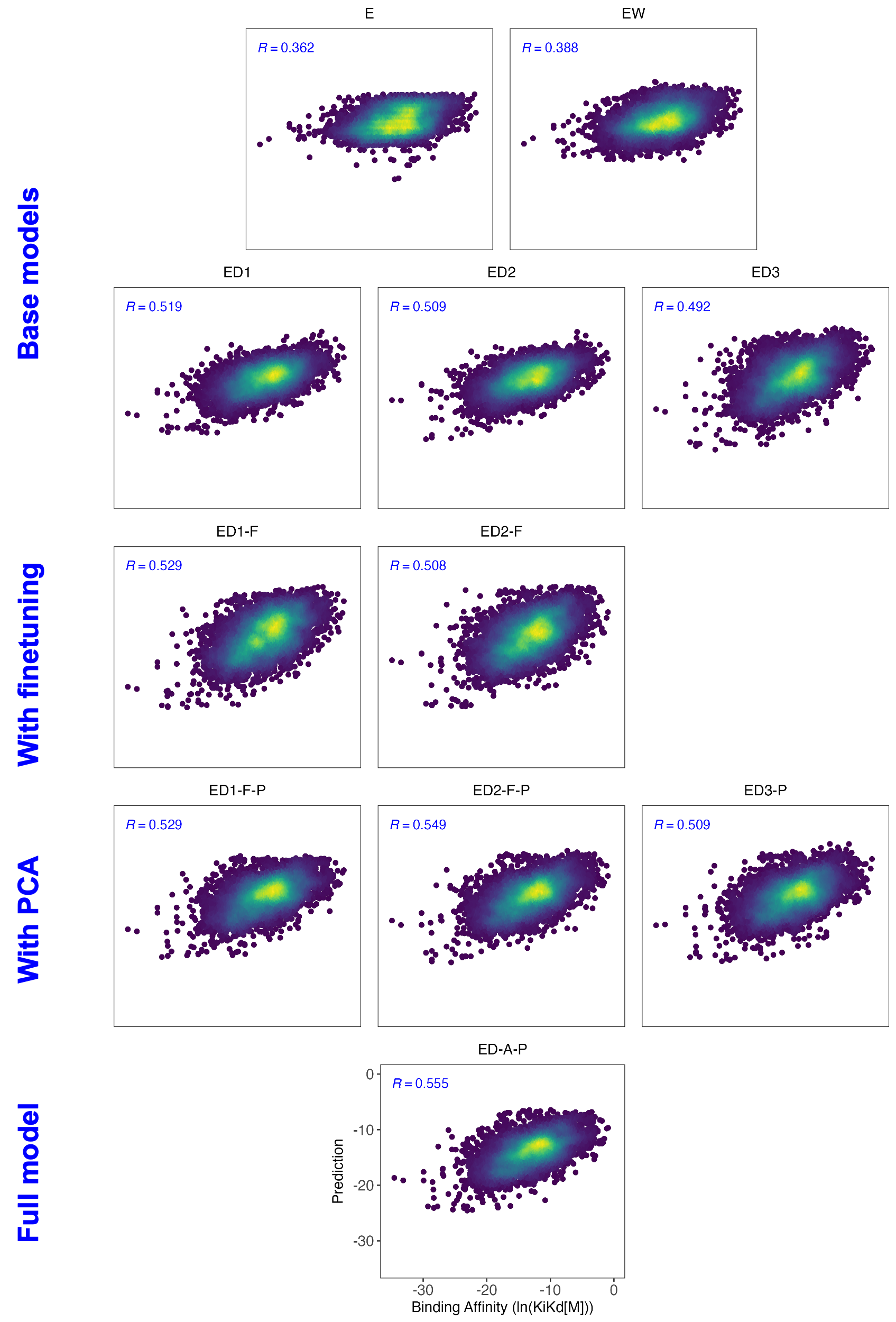}

\end{document}